%% file: lin_169-supp.tex
\documentclass[accepted]{uai2021}

\usepackage[american]{babel}

\usepackage{natbib} 
    \setcitestyle{authoryear,open={(},close={)}}
    
\usepackage{mathtools} 
\usepackage{booktabs} 
\usepackage{tikz} 

\usepackage{url}
\usepackage{graphicx}  
\usepackage{amsmath,amsfonts}
\usepackage{bm}
\usepackage{amssymb}
\usepackage{enumitem}

\usepackage{nicefrac}       
\usepackage{microtype}      
\usepackage{amsmath}
\usepackage{ntheorem}
\usepackage{multirow}
\usepackage{cleveref}
\usepackage{color}
\usepackage{comment}
\usepackage{algorithm}
\usepackage{algorithmic}
\usepackage[subrefformat=parens,labelformat=parens]{subfig}
\usepackage{physics}
\usepackage[title]{appendix}
\newtheorem{definition}{Definition}
\newtheorem{lemma}{Lemma}
\newtheorem{remark}{Remark}
\newtheorem{corollary}{Corollary}

\newtheorem{prop}{Proposition}
\newtheorem*{propstar}{Proposition}

\newtheorem*{myproof}{Proof}
\newcommand{\proofname}{Proof}
\DeclareMathOperator*{\argmax}{argmax} 
\DeclareMathOperator*{\argmin}{argmin} 
\DeclareMathOperator{\E}{\mathbb{E}}

\DeclareMathOperator{\b0}{\mathbf{0}}
\DeclareMathOperator{\bR}{\mathbb{R}}
\DeclareMathOperator{\bG}{\mathbf{G}}
\DeclareMathOperator{\bU}{\mathbf{U}}
\DeclareMathOperator{\bP}{\mathbf{P}}
\DeclareMathOperator{\bW}{\mathbf{W}}

\DeclareMathOperator{\bI}{\mathbf{I}}
\DeclareMathOperator{\bH}{\mathbf{H}}

\DeclareMathOperator{\cS}{\mathcal{S}}
\DeclareMathOperator{\cA}{\mathcal{A}}
\DeclareMathOperator{\cC}{\mathcal{C}}
\DeclareMathOperator{\cG}{\mathcal{G}}
\DeclareMathOperator{\cL}{\mathcal{L}}

\DeclareMathOperator{\cX}{\mathcal{X}}
\DeclareMathOperator{\cD}{\mathcal{D}}
\DeclareMathOperator{\cB}{\mathcal{B}}

\DeclarePairedDelimiterX{\inp}[2]{\langle}{\rangle}{#1, #2}

\DeclareMathAlphabet{\mathbmit}{OML}{cmm}{b}{it}
\renewcommand{\vec}[1]{\mathbmit{#1}}
\newcommand*{\MYQED}{\null\nobreak\hfill\ensuremath{\square}}%
\let\matr\vec

\makeatletter
\@tfor\next:=abcdefghijklmnopqrstuvwxyxz\do
 {\begingroup\edef\x{\endgroup
    \noexpand\@namedef{v\next}{\noexpand\vec{\next}}%
  }\x}
\@tfor\next:=ABCDEFGHIJKLMNOPQRSTUVWXYZ\do
 {\begingroup\edef\x{\endgroup
    \noexpand\@namedef{m\next}{\noexpand\matr{\next}}%
  }\x}
\makeatother

\makeatletter
\newcommand{\openbox}{\leavevmode
  \hbox to.77778em{%
  \hfil\vrule
  \vbox to.675em{\hrule width.6em\vfil\hrule}%
  \vrule\hfil}}
\makeatletter
\newcommand{\printfnsymbol}[1]{%
  \textsuperscript{\@fnsymbol{#1}}%
}

\newcommand*\xbar[1]{%
   \hbox{%
     \vbox{%
       \hrule height 0.5pt 
       \kern0.5ex
       \hbox{%
         \kern-0.1em
         \ensuremath{#1}%
         \kern-0.1em
       }%
     }%
   }%
}
\newcount\colveccount
\newcommand*\colvec[1]{
        \global\colveccount#1
        \begin{pmatrix}
        \colvecnext
}
\def\colvecnext#1{
        #1
        \global\advance\colveccount-1
        \ifnum\colveccount>0
                \\
                \expandafter\colvecnext
        \else
                \end{pmatrix}
        \fi
}

\title{Escaping from Zero Gradient: Revisiting Action-Constrained Reinforcement Learning via Frank-Wolfe Policy Optimization}

\author[1*]{Jyun-Li Lin}
\author[12*]{Wei Hung}
\author[1*]{Shang-Hsuan Yang}
\author[1]{\href{mailto:<pinghsieh@nctu.edu.tw>?Subject=Escaping from Zero Gradient: Revisiting Action-Constrained Reinforcement Learning via Frank-Wolfe Policy Optimization}{Ping-Chun Hsieh}}
\author[3]{Xi Liu}
\affil[1]{%
    Department of Computer Science\\
    National Yang Ming Chiao Tung University\\
    Hsinchu, Taiwan
}
\affil[2]{%
    Research Center for Information Technology Innovation\\
    Academia Sinica\\
    Taipei, Taiwan
}
\affil[3]{%
    Applied Machine Learning, Facebook AI, Menlo Park, CA, USA
}

\affil[*]{%
    Equal Contribution
}
\begin{document}
\maketitle

\begin{abstract}
Action-constrained reinforcement learning (RL) is a widely-used approach in various real-world applications, such as scheduling in networked systems with resource constraints and control of a robot with kinematic constraints.
While the existing projection-based approaches ensure zero constraint violation, they could suffer from the zero-gradient problem due to the tight coupling of the policy gradient and the projection, which results in sample-inefficient training and slow convergence.
To tackle this issue, we propose a learning algorithm that decouples the action constraints from the policy parameter update by leveraging state-wise Frank-Wolfe and a regression-based policy update scheme.
Moreover, we show that the proposed algorithm enjoys convergence and policy improvement properties in the tabular case as well as generalizes the popular DDPG algorithm for action-constrained RL in the general case.
Through experiments, we demonstrate that the proposed algorithm significantly outperforms the benchmark methods on a variety of control tasks.
\end{abstract}

\input{01-intro}

\input{03-model}

\input{05-policy}

\input{07-exp}

\input{02-related}

\input{08-conclusion}

\begin{acknowledgements} 
This material is based upon work partially supported by the Ministry of Science and Technology of Taiwan under Contract No. MOST 108-2636-E-009-014 and Contract No. MOST 109-2636-E-009-012.
\end{acknowledgements}

\bibliography{lin_169-supp}

\input{appendix}

\end{document}

%% file: 01-intro.tex
\section{Introduction}
\label{section:intro}


Action-constrained reinforcement learning (RL) is a popular approach for sequential decision making in real-world systems.
One classic example is maximizing the network-wide utility by optimally allocating the network resource under capacity constraints \citep{xu2018experience,gu2019intelligent,zhang2020cfr}.
Another example is robot control under kinematic constraints \citep{pham2018optlayer,gu2017deep,jaillet2012path,tsounis2020deepgait}, which capture the limitations of the physical components of a robot (e.g., in terms of velocity, torque, or output power).
In these examples, the constraints essentially characterize the set of feasible actions at each state.
To ensure the safe and normal operation of these real-world systems, it is required that these action constraints are satisfied throughout the evaluation as well as the training processes \citep{chow2018lyapunov,liu2020robust,gu2017deep}.
Therefore, in action-constrained RL, an effective training algorithm is required to achieve the following two tasks simultaneously: (i) iteratively improving the policy and (ii) ensuring zero constraint violation at each training step.


To enable RL with action constraints, one popular generic approach is to include an additional differentiable projection layer at the output of the policy network and follow the standard end-to-end policy gradient approach \citep{pham2018optlayer,dalal2018safe,bhatia2019resource}.  
While being a general-purpose solution, this projection layer could result in the \textit{zero-gradient issue} during training due to the tight coupling of the policy gradient update and the projection layer.
Specifically, zero gradient occurs when the original output of the policy network falls outside of the feasible action set and any small perturbation of the policy parameters does not lead to any change in the final output action due to the projection mechanism.
To better understand the zero-gradient issue, let us consider a toy example of a policy network with one hidden layer and a linear output layer used to produce a deterministic scalar action. Suppose the actions are required to be non-negative.
To satisfy the non-negativity action constraint, an additional $L_2$-projection layer, which is equivalent to a Rectified Linear Unit (ReLU), is added to the output of the policy.
It can be seen that the policy network can easily suffer from zero gradient due to the clipping effect of ReLU \citep{maas2013rectifier}.
If the zero-gradient issue occurs in a large portion of the state space, the training process could be sample-inefficient as most of the samples are wasted, and therefore the convergence speed could be slow.
Notably, the zero-gradient issue can be particularly severe in the early training phase since the pre-projection actions produced by the policy network are likely to be far away from the feasible sets.

The fundamental cause of the zero-gradient issue is the tight coupling of the policy parameter update and the projection layer under the standard policy gradient framework.
Specifically, in the end-to-end policy-gradient-based training process, the update of the policy parameters relies on the gradient of the actual policy output with respect to the policy parameters and thereby involves the gradient of this additional projection layer.
To escape from the zero-gradient issue, we take a different approach and propose a learning algorithm that \textit{decouples} the parameter update for policy improvement from constraint satisfaction, without using the policy gradient theorem.
The proposed algorithm can be highlighted as follows: 
\vspace{-1mm}
\begin{itemize}[leftmargin=*]
    \item To accommodate the action constraints, we leverage the Frank-Wolfe method \citep{frank1956algorithm} to search for feasible action update directions directly \textit{within} the feasible action sets in a \textit{state-wise} manner. 
    Through this procedure, for a collection of states, we obtain the reference actions that are used to guide the update of the policy parameters for improving the current policy.
    \vspace{-1mm}
    \item To update the parameters of the policy network, we propose to construct a loss function (e.g., mean squared error) that enables the policy network to adjust its outputs toward the reference actions.
    This update scheme can be viewed as solving a regression problem based on the reference actions by taking one-step gradient descent.
    In this way, the parameter update is completely decoupled from the action constraints.
\end{itemize}
\vspace{-2mm}

Since the proposed framework obviates the need for the gradient of a projection layer, it avoids the zero-gradient issue by nature. 

\vspace{-1mm}

\textbf{Our Contributions.} 
In this paper, we revisit the action-constrained RL problem and propose a novel learning framework that avoids the zero-gradient issue and achieves zero constraint violation simultaneously:
\vspace{-2mm}

\begin{itemize}[leftmargin=*]
    \item We formally identify the important zero gradient issue in the existing projection-based approaches for action-constrained RL.
    We also pinpoint that the fundamental cause of the  zero-gradient issue is the tight coupling of the policy parameter update and the projection layer under the standard policy gradient framework.
    To the best of our knowledge, this is the first time that the zero-gradient issue is discussed in the context of action-constrained RL.
    \item To better describe the proposed learning framework, we start from the case of finite state spaces and introduce Frank-Wolfe policy optimization (FWPO) with tabular policy parameterization, which can be viewed as an instance of the generalized policy iteration.
    By directly searching for update directions within the feasible sets via state-wise Frank-Wolfe, FWPO automatically achieves zero constraint violation and does not require any additional projection.
    Moreover, we establish the convergence of FWPO as well as its policy improvement property.
    \vspace{-1mm}
    \item Built on FWPO, we propose Neural FWPO (NFWPO) by extending the idea of FWPO to the general neural policies via a regression argument.
    By constructing a loss function and leveraging state-wise Frank-Wolfe, we decouple the policy parameter update from the action constraints.
    This design automatically prevents the zero-gradient issue. 
    Moreover, we show that the vanilla DDPG is a special case of NFWPO if there is no action constraints.
    \vspace{-1mm}
    \item Through experiments on various real applications, we empirically show the zero-gradient problem and demonstrate that the proposed algorithms significantly outperform the popular benchmark methods for action-constrained RL.
\end{itemize}









%% file: 03-model.tex
\section{Preliminaries}
\label{section:model}
We consider an infinite-horizon discounted Markov decision process (MDP) defined by a tuple $(\mathcal{S},\mathcal{A},p,r,\gamma)$, where $\mathcal{S}$ is state space, $\mathcal{A}$ denotes the action space, $p$ is the state transition probability, $r$ is the reward function, and $\gamma\in (0,1)$ denotes the discount factor.
We assume that the action space $\mathcal{A}\subseteq \bR^{N}$ is continuous and the reward function takes value in $[0,1]$ for all state-action pairs.
At each time step $t=0,1,\cdots$, the learner observes state $s_t$, takes an action $a_t$, and receives an immediate reward $r_t$.
In this paper, we consider the \textit{action-constrained} MDPs where for each state $s\in \cS$ there is a feasible action set $\cC(s) \subseteq \cA$ determined by the underlying collection of constraints.
We assume that $\cC(s)$ is compact and convex.
In this paper, we focus on deterministic policies and use $\pi(\cdot;\theta):\cS\rightarrow \cA$ to denote a deterministic parametric policy with parameter vector $\theta\in\bR^n$. 
Under a policy $\pi$, the value functions are defined as the expected long-term rewards 
\begin{align}
    V(s;\pi)&=\E\Big[\sum_{t=0}^{\infty} \gamma^t r(s_t,a_t)\rvert s_0=s,\pi\Big],\\
    Q(s,a;\pi)&=\E\Big[\sum_{t=0}^{\infty} \gamma^t r(s_t,a_t)\rvert s_0=s,a_0=a,\pi\Big].
\end{align}
To make a comparison between policies, for any two policies $\pi$ and $\pi'$, we say that $\pi\geq \pi'$ if $V(s;\pi)\geq V(s;\pi')$, for all $s\in\cS$.
This essentially constructs a partial ordering among policies.
To construct a total ordering of all the policies, consider the performance objective defined as a weighted average of the value function
\begin{equation}
 J_{\mu}(\pi):=\E_{s\sim \mu}[V(s;\pi)],  
\end{equation}
where $\mu$ is called the restarting state distribution \citep{kakade2002approximately}.
Note that one common choice of $\mu$ is the initial state distribution.
It is also convenient to define the discounted state visitation distribution $d_\mu^{\pi}$ as $d_\mu^{\pi}(s):=(1-\gamma)\E_{s_0\sim \mu}[\sum_{t=0}^{\infty}\gamma^t P(s_t=s \rvert s_0,\pi)]$, for each $s\in \cS$.

\textbf{Notations}. We use the standard notations $\norm{\cdot}_p$ and $\norm{\cdot}_{F}$ to denote the $L_p$-norm of a vector and the Frobenius norm of a matrix, respectively. We use $\inp{\cdot}{\cdot}$ to denote the inner product of two real vectors. For a set $\cD$, we define the diameter of $\cD$ as $\text{diam}_{\norm{\cdot}_2}(\cD):=\sup_{x_1,x_2\in \cD}\norm{x_1-x_2}_2$.
We use $\text{dom}f$ to denote the domain of a function $f$.

\subsection{Policy Gradient}
\label{section:model:DPG}
To optimize the objective $J_{\mu}(\pi)$, the typical approach is to apply gradient ascent based on the policy gradient.
Under the standard regularity conditions, the deterministic policy gradient \citep{silver2014deterministic} can be written as
\begin{align}
    &\nabla_{\theta} J_{\mu}(\pi(\cdot;\theta))\nonumber\\
    &= \E_{s\sim d_{\mu}^{\pi}}\Big[\nabla_{\theta}\pi(s;\theta) \nabla_{a}Q(s,a;\pi(\cdot;\theta))\rvert_{a=\pi(s;\theta)}\Big].
    \label{eq:DPG}
\end{align}

As a practical implementation of the deterministic policy gradient approach, DDPG \citep{lillicrap2016continuous} extends deep $Q$-learning \citep{mnih2015human} to continuous action space in an actor-critic manner.
Specifically, DDPG updates the policy parameter $\theta$ by applying stochastic gradient ascent according to (\ref{eq:DPG}) and obtains an approximated $Q$-function $Q(s,a; \phi)$ parameterized by $\phi$ by using a $Q$-learning-like critic, which updates $\phi$ by minimizing the loss $\E_{(s,a,s',r)\sim \rho}[(r+\gamma Q(s',\pi(s';\theta^-);\phi^-)-Q(s,a;\phi))^2]$, where $\rho$ denotes the sampling distribution of the replay buffer, $\theta^-$ and $\phi^-$ are the parameters of the actor and critic target networks, respectively.





\subsection{Frank-Wolfe Methods}
In this section, we provide an overview of the Frank-Wolfe algorithms. Consider an optimization problem in the form
\begin{equation}
    \max_{{x}\in\mathcal{X}} F({x}),\label{eq:constrained opt}
\end{equation}
where $F(\cdot):\mathbb{R}^{d}\rightarrow \mathbb{R}$ is a differentiable function with a Lipschitz continuous gradient, $\mathcal{X}\subseteq \bR^d$ is the feasible set characterized by the underlying constraints on ${x}$.
One popular approach is to apply the projected gradient ascent method \citep{bubeck2015convex}, which combines the standard gradient ascent with a projection step.
By contrast, as a projection-free method, the classic Frank-Wolfe algorithms \citep{frank1956algorithm} and its variants solve the constrained optimization problems in (\ref{eq:constrained opt}) by leveraging a first-order subproblem.
We briefly summarize the Frank-Wolfe algorithm for non-convex objective functions in the batch settings as follows \citep{lacoste2016convergence,reddi2016stochastic}:

\vspace{-2mm}
\begin{itemize}[leftmargin=*]
    \item \textbf{Initialization.} Let $x_k$ denote the input at the $k$-th iteration and choose an arbitrary $x_0\in\cX$ to be the initial point. 
    \vspace{-1mm}
    \item \textbf{Search for an update direction within the feasible set.} 
    In the $k$-th iteration, compute $v_k=\argmax_{v\in\cX}\inp{v}{\nabla_{x}F(x)\rvert_{x=x_k}}$ and update the iterate as $x_{k+1}=x_k+\beta_k(v_k-x_k)$, where $v_k-x_k$ is the update direction and $\beta_k$ denotes the learning rate.
\end{itemize}

For unconstrained optimization problems, the convergence properties are typically analyzed in terms of the gradient norm $\norm{\nabla_x F(x)}_2$.
By contrast, for constrained maximization problems, one widely-used metric of convergence in the Frank-Wolfe literature is the \textit{Frank-Wolfe gap} defined as $\cG(x):=\max_{z\in \cX}\inp{z-x}{\nabla_{x}F(x)}$\footnote{In the literature, the Frank-Wolfe gap is typically defined as $\max_{z\in \cX}\inp{z-x}{-\nabla_{x}F(x)}$ since the goal is to minimize an objective function. By contrast, as the goal of RL is to optimize the policy in terms of rewards, we consider the maximization problem in the form of (\ref{eq:constrained opt}) and make the required changes accordingly.}.
It is easy to verify that $\cG(x)=0$ is a necessary and sufficient condition of that $x$ is a stationary point. 

%% file: 05-policy.tex
\section{Frank-Wolfe Policy Optimization}
\label{section:alg}

In this section, we formally present the proposed learning algorithms for action constrained RL.
To better describe the proposed learning framework, we start from a stylized setting with tabular policy parameterization for finite state spaces and extend the idea to develop a more practical algorithm for the general parametric policies.

\subsection{Frank-Wolfe Policy Optimization With Direct Policy Parameterization (FWPO)}
For ease of exposition, we first illustrate the proposed algorithm for the case of finite state spaces and tabular policies with direct parameterization, i.e., $\pi(s;{\theta})\equiv\theta(s)$, for all $s\in\mathcal{S}$.
We consider the performance objective $J_{\mu}(\pi)$ with some restarting state distribution $\mu$ with $\mu(s)>0$, for all $s\in\cS$, and define $\mu_{\min}:=\min_{s\in\cS}\mu(s)$.
For ease of notation, we also define $D_s:=\text{diam}_{\norm{\cdot}_2}(\cC(s))$ for each $s$ and $D_{\max}:=\max_{s\in\cS}D_s$.

Now we present the proposed FWPO algorithm. We use $\theta_k$ to denote the policy parameters in the $k$-th iteration and choose feasible initial policy parameters $\theta_0$ which satisfy $\theta_0(s)\in \cC(s)$, for all $s\in\cS$. FWPO adopts the generalized policy iteration framework \citep{sutton2018reinforcement} by alternating between two subroutines in each iteration: 
\vspace{-2mm}

\begin{itemize}[leftmargin=*]
    \item \textbf{Policy update via state-wise Frank-Wolfe.}
FWPO updates the policy by finding a feasible update direction of each state $s\in\cS$ via Frank-Wolfe as
\begin{align}
    \hspace{-6pt}c_{k}(s)&=\argmax_{c \in \cC(s)}~\inp{c}{\nabla_{a} Q(s,a;\pi(\cdot;\theta_k))\rvert_{a=\theta_k(s)}},\label{eq:FW direction in tabular case}\\
    \hspace{-6pt}\theta_{k+1}(s)&=\theta_k(s)+\alpha_k(s) (c_k(s)-\theta_k(s)),\label{eq:FW update in tabular case}
\end{align}
where $c_k(s)-\theta_k(s)$ is the update direction and $\alpha_k(s)$ denotes the (state-dependent) learning rate.
Moreover, it is natural to define the \textit{state-wise Frank-Wolfe gap} of the $Q$-function at $\theta_k$ as 
\begin{equation}
    g_{k}(s):=\inp{c_k(s)-\theta_k(s)}{\nabla_{a} Q(s,a;\pi(\cdot;\theta_k))\rvert_{a=\theta_k(s)}}.\label{eq:state-wise FW gap}
\end{equation}
It is easy to verify that $g_{k}(s)\geq 0$, for all $k\in\mathbb{N}$ and for all $s\in\cS$.
As will be shown momentarily, to ensure convergence, the learning rate is configured to be $\alpha_k(s)=\frac{(1-\gamma)\mu_{\min}}{L D_s^2}g_k(s)$.
\item \textbf{Evaluation of the current policy.} FWPO then evaluates the updated policy and obtain the $Q$-function (or an approximated version) for the next iteration. This can be done by a standard policy evaluation approach.
\end{itemize}

The above scheme of FWPO is detailed in Algorithm \ref{alg:FWPO}.
As suggested by Algorithm \ref{alg:FWPO}, FWPO always searches for an update direction within the feasible action sets. Therefore, FWPO automatically achieves zero constraint violation and does not require any additional projection by nature.

\begin{algorithm}[!htbp]
   \caption{Frank-Wolfe Policy Optimization (FWPO)}
   \label{alg:FWPO}
\begin{algorithmic}[1]
   \STATE {\bfseries Input:} Initialize the policy parameters as $\theta_0$ that satisfies $\theta_0(s)\in \cC(s)$ for all $s\in\cS$
    \FOR{each iteration $k=0,1,\cdots$}
      \STATE Evaluate $\pi(\cdot;\theta_k)$ and obtain $Q(s,a;\pi(\cdot;\theta_k))$
      \FOR{each state $s\in\cS$}
      \STATE Compute the Frank-Wolfe update direction by $c_{k}(s)=\argmax_{c \in \cC(s)}\inp{c}{\nabla_{a} Q(s,a;\pi(\cdot;\theta_k))}$
      \STATE $g_{k}(s)=\inp{c_k(s)-\theta_k(s)}{\nabla_{a} Q(s,a;\pi(\cdot;\theta_k))}$
      \STATE $\alpha_k(s)=\frac{(1-\gamma)\mu_{\min}}{L D_s^2}g_k(s)$
      \STATE $\theta_{k+1}(s)=\theta_k(s)+\alpha_k(s) (c_k(s)-\theta_k(s))$
      \ENDFOR
    \ENDFOR
\end{algorithmic}
\end{algorithm}

\begin{remark}
\label{remark:FWPO}
\normalfont One salient feature of FWPO is that the policy update in (\ref{eq:FW direction in tabular case})-(\ref{eq:FW update in tabular case}) is done by searching for feasible update directions based on $\nabla_a Q(s,a;\pi)$ on a \textit{per-state} basis with state-dependent learning rates, instead of using the standard policy gradient of the performance objective $J_{\mu}(\pi)$. 
As will be seen in Section \ref{section:alg:NFWPO}, this design plays a critical role in decoupling the policy parameter update from constraint satisfaction.
Another advantage of FWPO is that it is agnostic to the discounted state visitation distribution $d_{\mu}^{\pi}$ (\textit{cf.} the deterministic policy gradient in (\ref{eq:DPG})) due to the state-wise nature.
This feature allows FWPO to be directly applicable in the off-policy settings in its original form\footnote{In the off-policy settings, the deep policy gradient approaches typically require dropping a term in the policy gradient expression to accommodate the behavior policy \citep{silver2014deterministic}.}.
\end{remark}
\begin{remark}
\label{remark:FWPO convergence}
\normalfont As the policy update under FWPO is done on a state-by-state basis instead of directly on $J_{\mu}(\pi)$, the convergence guarantees of the standard Frank-Wolfe methods do not directly apply to the objective $J_{\mu}(\pi)$ under FWPO.
From this perspective, FWPO is \textit{not} a trivial combination of the Frank-Wolfe methods and policy iteration.
\end{remark}

\vspace{-3mm}
As suggested by Remark \ref{remark:FWPO convergence}, we proceed to establish the convergence result of FWPO.
For the convergence analysis, based on the state-wise Frank-Wolfe gaps defined in (\ref{eq:state-wise FW gap}), we define the \textit{effective Frank-Wolfe gap} of $J_{\mu}(\pi(\cdot;\theta))$ at $\theta_k$ as 
\begin{equation}
    \cG_k:=\Big(\sum_{s\in\cS} g_k(s)^2\Big)^{1/2}.\label{eq:effective FW gap}
\end{equation}
Note that $\cG_k=0$ if and only if the update direction is zero for all the states, i.e., $c_k(s)-\theta_k(s)=0$.
Hence, $\cG_k$ indicates whether the $J_{\mu}(\pi(\cdot;\theta))$ converges to a stationary point.
We also define $\bar{\cG}_T:=\min_{0\leq k\leq T}\cG_k$.
To establish the convergence results, we also assume mild regularity conditions on $r$ and $p$ as follows.

\vspace{-1mm}
\begin{definition}
\label{def:L smoothness}
\normalfont A differentiable function $f:\text{dom}f \rightarrow \bR$ is said to be \textit{$L_0$-smooth} if there exists $L_0\geq 0$ such that for any $x,y\in \text{dom}f$, $\norm{\nabla f(x) - \nabla f(y)}_2\leq L_0\norm{x-y}_2$.
\end{definition}

\vspace{-2mm}
\textbf{Regularity Assumptions:}
\vspace{-1mm}

\noindent \textbf{(A1)} The reward function $r(s,a)$ is differentiable and is $L_r$-smooth in $a$, for all $s,a$.
\vspace{-1mm}

\noindent \textbf{(A2)} The transition probability $p(s'\rvert s,a)$ is twice differentiable and $L_p$-smooth in $a$, for all $s,s',a$. Moreover, $p(s'\rvert s,a)$ satisfies $\sup_{s,a,s'}\norm{\nabla_{a}p(s'\rvert s,a)}_{2}< C_p$.

As the first step, we introduce the following proposition on the smoothness of the performance objective $J_\mu(\pi(\cdot; \theta))$.
Notably, given the regularity assumptions of $r$ and $p$ in action, it remains non-trivial to establish the smoothness of $J_{\mu}(\pi(\cdot; \theta))$ in $\theta$ due to the multi-step compound effect of the changes in policy parameters on the value functions.

\vspace{-1mm}
\begin{prop}
\label{PROP: L-SMOOTH OF J}
Under the regularity assumptions (A1)-(A2), there exists some constant $L>0$ such that for any restarting state distribution $\mu$, $J_\mu(\pi(\cdot;\theta))$ is $L$-smooth in $\theta$.
\end{prop}
\vspace{-3mm}

The proof of Proposition \ref{PROP: L-SMOOTH OF J} is provided in Appendix A.1. 
Now we are ready to present the convergence result.

\vspace{-2mm}
\begin{prop}
\label{PROP:CONVERGENCE}
Under the FWPO algorithm with $\alpha_k(s)=\frac{(1-\gamma)\mu_{\min}}{L D_s^2}g_k(s)$, $\{\pi(\cdot;\theta_k)\}$ form a non-decreasing sequence of policies in the sense that $\pi(\cdot;\theta_{k+1})\geq \pi(\cdot;\theta_k)$, for all $k$.
Moreover, the effective Frank-Wolfe gap of FWPO converges to zero as $k\rightarrow \infty$, and the convergence rate can be quantified as
\begin{equation}
    \sum_{k=0}^{\infty} \cG_k^2\leq \frac{2LD_{\max}^2}{(1-\gamma)^3 \mu_{\min}^2},
\end{equation}
which implies that $\bar{\cG}_T=O(T^{-1/2})$.
\end{prop}
\vspace{-3mm}

\begin{myproof}
\normalfont Due to space limitation, we provide a sketch of proof:
(i) To show the non-decreasing property, we leverage the policy difference lemma \citep{kakade2002approximately} and verify a sufficient condition of strict policy improvement; (ii) To show the convergence result, we leverage the smoothness of the value functions as well as the objective and use the technique for convergence of non-convex optimization similar to that in \citep{reddi2016stochastic,lacoste2016convergence}; (iii) A proper learning rate can be selected by taking the smoothness conditions as well as the restarting state distribution into account. 
For completeness, the detailed proof is provided in Appendix A.2. 
\end{myproof}
\vspace{-3mm}

\begin{remark}
\normalfont The style of the convergence guarantee in Proposition \ref{PROP:CONVERGENCE} is common in the analysis of gradient descent methods for non-convex smooth functions \citep{bottou2018optimization}.
Moreover, the result (i.e., convergence to a stationary point) in Proposition \ref{PROP:CONVERGENCE} resembles those of the policy gradient algorithms \citep{sutton2000policy,silver2014deterministic}, but for the action-constrained RL settings.
On the other hand, in (\ref{eq:FW direction in tabular case}), the search of the update direction requires the gradient of the $Q$-function.
In practice, it may not be feasible to obtain the whole true $Q$-function, and a value function approximator can be included. In practice, it can be expected that a sufficiently accurate critic shall provide a sufficiently good update direction.
\end{remark}
\vspace{-2mm}

\vspace{-2mm}
\subsection{Neural Frank-Wolfe Policy Optimization (NFWPO)}
\label{section:alg:NFWPO}
\vspace{-2mm}

In this section, we formally present the
proposed NFWPO algorithm for general parametric policies for action-constrained RL.
As highlighted in Section \ref{section:intro}, we propose to decouple constraint satisfaction from the policy parameter update.
Specifically, to accommodate the action constraints, we extend the state-wise Frank-Wolfe subroutine to the general parametric policies.
One inherent challenge of such extension is that the Frank-Wolfe method searches for an update direction within the feasible set by nature.
However, under neural parameterization, an action produced by the neural network is not guaranteed to stay in the feasible action set. 
To address this, we propose to incorporate a projection step into the state-wise Frank-Wolfe subroutine.
Define a projection operator as
\begin{equation}
    \Pi_{\cC(s)}(z)=\argmin_{y\in \cC(s)}\norm{y-z}_2.
\end{equation}
For ease of exposition, in the sequel we call the input $z$ a \textit{pre-projection action} and $\Pi_{\cC(s)}(z)$ a \textit{post-projection action}.

NFWPO adopts the actor-critic architecture.
Let $\bar{\theta}$ and $\bar{\phi}$ be the current parameters of the actor and the critic, respectively. 
The main features of NFWPO are captured by the actor part as below.
\vspace{-2mm}

\begin{itemize}[leftmargin=*]
    \item \textbf{Derive reference actions via state-wise Frank-Wolfe.} For each $s$ in the mini-batch $\cB$, NFWPO uses Frank-Wolfe to compute the reference action at each state $s$ as
    \begin{equation}
        \tilde{a}_s=\Pi_{\cC(s)}(\pi(s;\bar{\theta}))+{\alpha} \big(\bar{c}(s)-\Pi_{\cC(s)}(\pi(s;\bar{\theta}))\big),\label{eq:FW target action}
    \end{equation}
    where $\alpha$ is the learning rate of Frank-Wolfe and
    \begin{align}
        \bar{c}(s)&=\argmax_{c \in \cC(s)}\inp{c}{\nabla_{a} Q(s,a;\bar{\phi})\rvert_{a=\Pi_{\cC(s)}(\pi(s;\bar{\theta}))}}.\label{eq:FW direction}
    \end{align}
    (Note that the projection $\Pi_{\cC(s)}(\cdot)$ is only for generating feasible actions and does not require backpropagation.)
    \vspace{-1mm}
    \item \textbf{Construct an MSE loss function.} NFWPO constructs a loss function $L_{\text{NFWPO}}(\theta;\bar{\theta})$ as the MSE between the actions of the current policy and the reference actions, i.e.,
    \begin{align}
        \cL_{\text{NFWPO}}(\theta;\bar{\theta})=\sum_{s\in\cB}\big(\pi(s;\theta)- \tilde{a}_s\big)^2.\label{eq:NFWPO MSE loss}
    \end{align}
    \vspace{-1mm}
    \item \textbf{Update policy by gradient descent.} NFWPO updates the policy parameter by minimizing the MSE loss in (\ref{eq:NFWPO MSE loss}) by using gradient descent for one step, i.e.,
    \begin{equation}
        \theta\leftarrow\theta - {\beta}\nabla_\theta \cL_{\text{NFWPO}}(\theta;\bar{\theta}).\label{eq:NFWPO MSE update}
    \end{equation}
\end{itemize}
\vspace{-1mm}

On the other hand, the critic of NFWPO can be based on any standard policy evaluation technique. For ease of exposition, for NFWPO, we use the same critic as the vanilla DDPG (as described in Section \ref{section:model:DPG}). The detailed pseudo code of NFWPO is provided in the supplementary material.

Notably, similar to (\ref{eq:FW direction in tabular case}), NFWPO only uses $\nabla_a Q(s,a;\bar{\phi})$ for deriving reference actions, \textit{without} using the deterministic policy gradient in (\ref{eq:DPG}).
This design allows NFWPO to decouple constraint satisfaction in (\ref{eq:FW target action})-(\ref{eq:FW direction}) from the parameter update in (\ref{eq:NFWPO MSE loss})-(\ref{eq:NFWPO MSE update}).
As highlighed in Section \ref{section:intro}, this decoupling obviates the need for the gradient of a projection layer and hence automatically avoids the zero-gradient issue.
Moreover, below we show that DDPG is actually a special case of NFWPO when there is no action constraints. The proof is provided in Appendix B 

\vspace{-1mm}
\begin{prop}
\label{PROP:NFWPO DDPG}
If there is no action constraints, then the policy update scheme of NFWPO in (\ref{eq:FW target action})-(\ref{eq:NFWPO MSE update}) is equivalent to the vanilla DDPG by \citep{lillicrap2016continuous}.
\end{prop}

\vspace{-3mm}
\begin{remark}
\normalfont While NFWPO leverages a projection step in (\ref{eq:FW target action}), this projection step is only for deriving reference actions and does not take part in the policy parameter update.
As a result, NFWPO does not require backpropagation of the projection step (as shown in (\ref{eq:FW target action})-(\ref{eq:NFWPO MSE update})) and therefore automatically avoids the zero-gradient issue.
Hence, NFWPO is essentially different from the existing solutions that combine DDPG with a projection layer for end-to-end training \citep{pham2018optlayer,dalal2018safe}.
\end{remark}
\vspace{-2mm}

%% file: 07-exp.tex
\vspace{-3mm}
\section{Experimental Results}
\label{section:exp}
\vspace{-1mm}

In this section, we empirically evaluate FWPO and NFWPO in various real-world applications, including bike sharing systems, communication networks, and continuous control in MuJoCo.
We compare the proposed algorithms against the following popular benchmark methods:
\vspace{-2mm}

\begin{itemize}[leftmargin=*]
    \item \textbf{DDPG+Projection}: The training procedure is identical to the vanilla DDPG \citep{lillicrap2016continuous} except that the action is post-processed by the $L_2$-projection operator $\Pi_{\cC(s)}(\cdot)$ before being applied to the environment.
    \vspace{-0mm}
    \item \textbf{DDPG+RewardShaping}: Built on DDPG+Projection, this algorithm adds the $L_2$-norm between the pre-projection and post-projection actions as a penalty to the intrinsic reward.
    \vspace{-0mm}
    \item \textbf{DDPG+OptLayer}: This design uses a differentiable projection layer, namely the OptLayer, that supports end-to-end training via gradient descent \citep{pham2018optlayer}. 
\end{itemize}
\vspace{-2mm}

Moreover, for the projection step (without the need of backpropagation) required by DDPG+Projection, DDPG+RewardShaping, and NFWPO, we implement this functionality on the Gurobi optimization solver \citep{gurobi}.
Therefore, the post-projection actions are guaranteed to satisfy the action constraints for all the algorithms.
For each task, each algorithm is trained under the common set of 5 random seeds. Each evaluation consists of 10 episodes, and we report the average performance along with the standard deviation in Figures \ref{fig:BSS-3}-\ref{fig:Halfcheetah all}.
We also summarize the average return over the final 10 evaluations in Table \ref{tab:expResult} and Table \ref{tab:mujocoResult}. 
The detailed training setup can be found in Appendix D. The code of our experiments is available \footnote{https://github.com/upupsheep/NFWPO\_Final\_Code}.

\vspace{-1mm}
\subsection{\textbf{Bike Sharing Systems}}
\label{section:exp:BSS}
\vspace{-1mm}


We use the open-source BSS simulator\footnote{BSS: https://github.com/bhatiaabhinav/gym-BSS}, which was originally proposed by \citep{ghosh2017incentivizing} and later used for evaluating action-constrained RL by \citep{bhatia2019resource}.
%
In a bike-sharing problem, there are $m$ bikes and $n$ stations, each of which has a pre-determined bike storage capacity $C$. 
An action is to allocate $m$ bikes to $n$ stations under random demands.
The reward signal consists of three parts: (i) Moving cost: the cost of moving one or multiple bikes from one station to another; (ii) Lost-demand cost: the cost of unserved demand due to bike outage. (iii) Overflow cost: the cost incurred when the number of bikes in one station exceeds its capacity. 
\vspace{-1mm}

\begin{figure}[!htb]
\centering
$\begin{array}{c c}
    \multicolumn{1}{l}{\mbox{\bf }} & \multicolumn{1}{l}{\mbox{\bf }}   \\ 
    \hspace{-3mm} \scalebox{0.26}{\includegraphics[width=\textwidth]{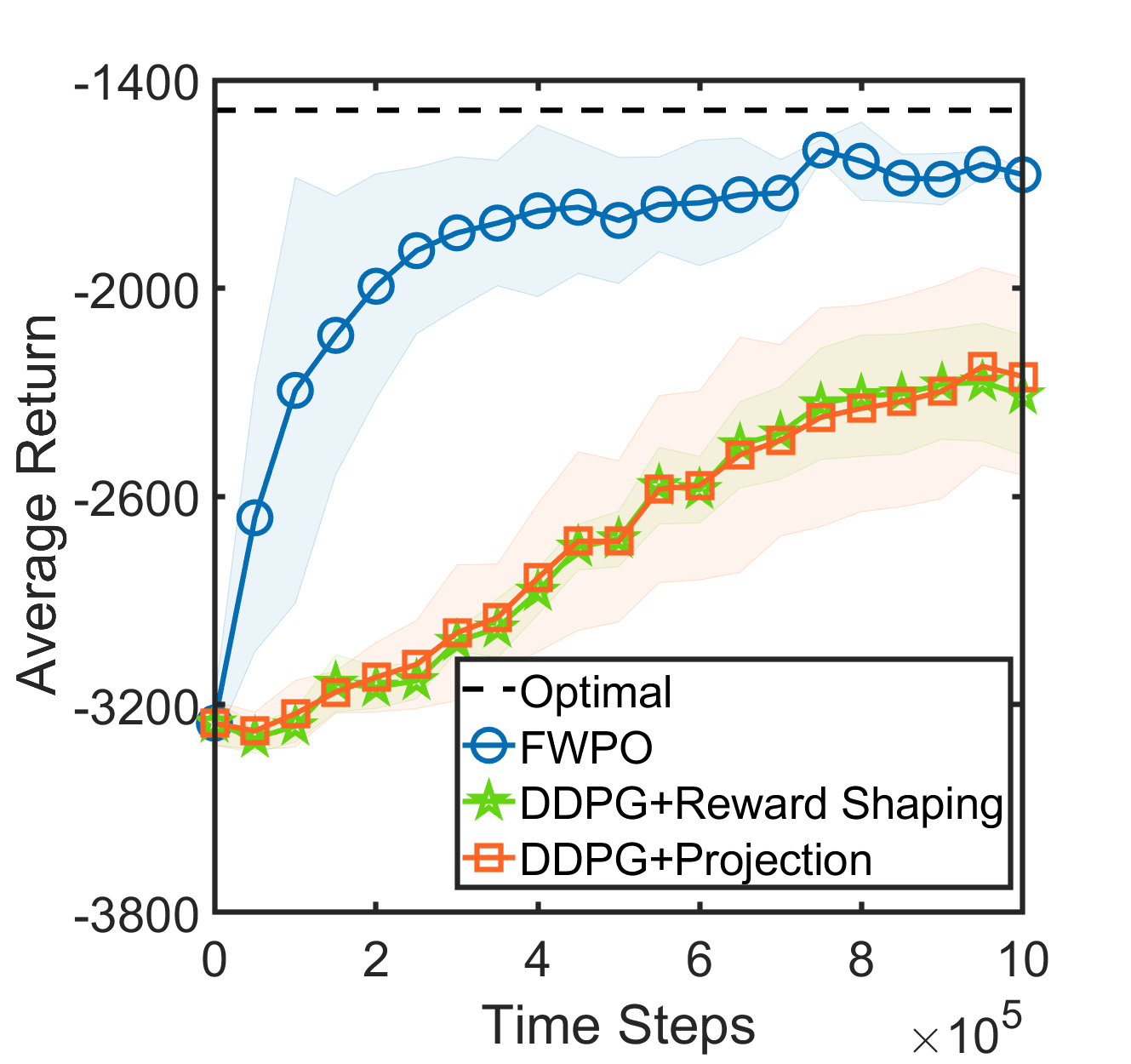}}  \label{fig:Bike_sharing_reward} & 
    \hspace{-6mm} \scalebox{0.26}{\includegraphics[width=\textwidth]{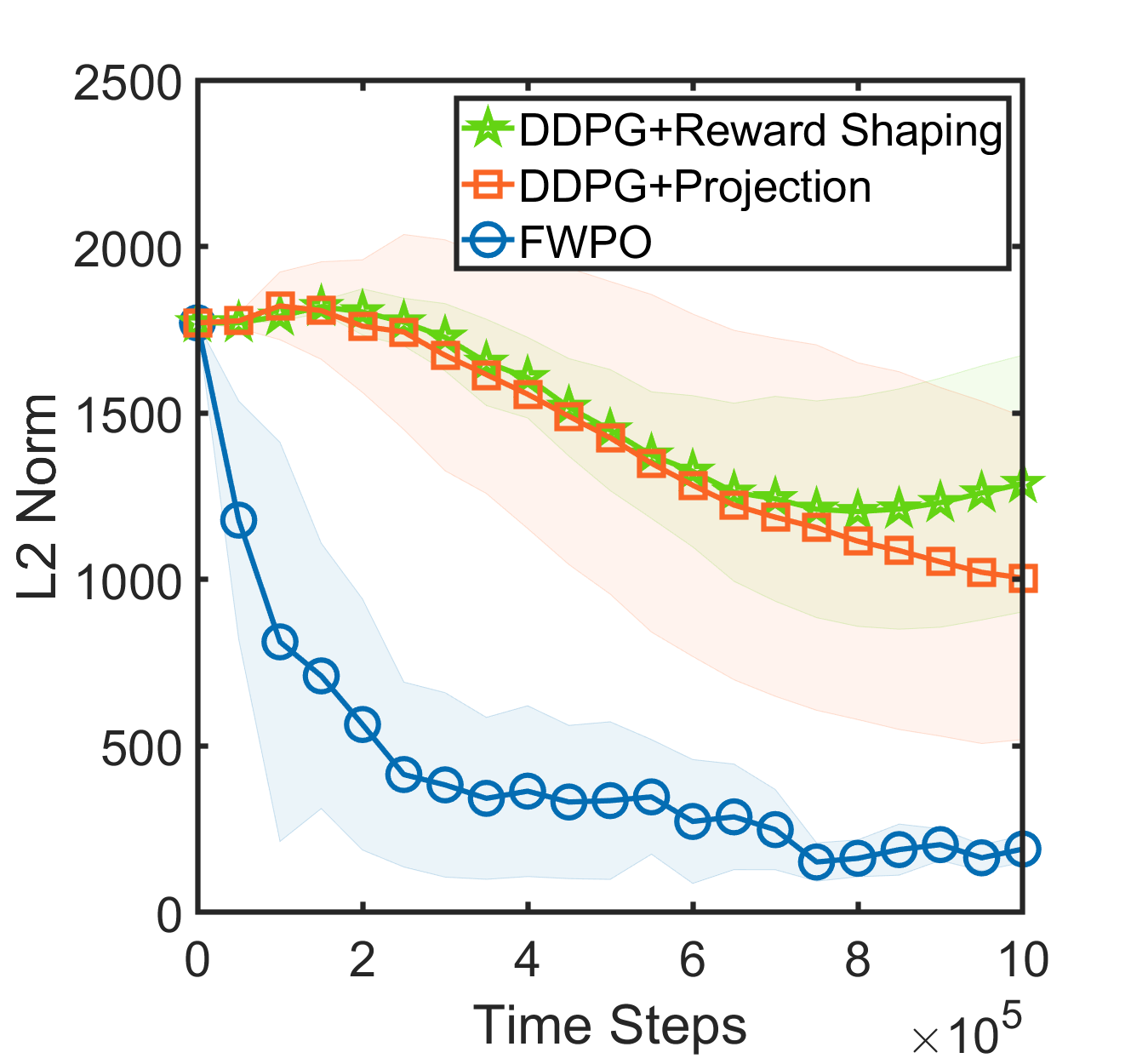}}
    \label{fig:Bike_sharing_norm} \\
    \mbox{\small(a)} & \hspace{-2mm} \mbox{\small(b)}
\end{array}$
\caption{Bike sharing problem with $n=3$ (BSS-3) under tabular policies: (a) Average return over 5 random seeds; (b) $L_2$-norm between the learned policies and the optimal policy at each training step.}
\label{fig:BSS-3}
\end{figure}

\textbf{Evaluating FWPO.} 
Since the bike sharing environment has a finite state space, we first use it to evaluate FWPO against the baseline methods, all with tabular policy parameterization.
For the action value function, we use the same $Q$-learning-like critic as the vanilla DDPG for all the algorithms.
A medium-sized system with $n=3$, $m=90$, and $C=35$ is chosen that allows to analytically find the optimal policy.
There are two types of constraints: (i) Global constraint: all the action entries shall sum to $90$; (ii) Local constraints: each entry of the action shall be between $0$ and $35$.
Figure \ref{fig:BSS-3}(a) shows the average return of the three algorithms. 
We observe that FWPO performs the best, while DDPG+Projection and DDPG+RewardShaping both suffer from slow learning.
This is also reflected by Figure \ref{fig:BSS-3}(b), which shows that FWPO converges to a near-optimal policy much faster than the baselines.
The above phenomenon is mainly due to the inaccurate policy gradient of DDPG under action constraints.
Specifically, the critics of the two baselines are trained with samples with feasible actions while the gradients $\nabla_a Q(s,a;\phi)$ are mostly evaluated at those actions outside the feasible sets.
By contrast, FWPO always stays in the feasible action sets and hence naturally avoids the issue of inaccurate gradients.

\begin{figure*}[!tb]
\centering
$\begin{array}{c c c}
    \multicolumn{1}{l}{\mbox{\bf }} & \multicolumn{1}{l}{\mbox{\bf }} & \multicolumn{1}{l}{\mbox{\bf }} \\ 
    \hspace{-0mm} \scalebox{0.255}{\includegraphics[width=\textwidth]{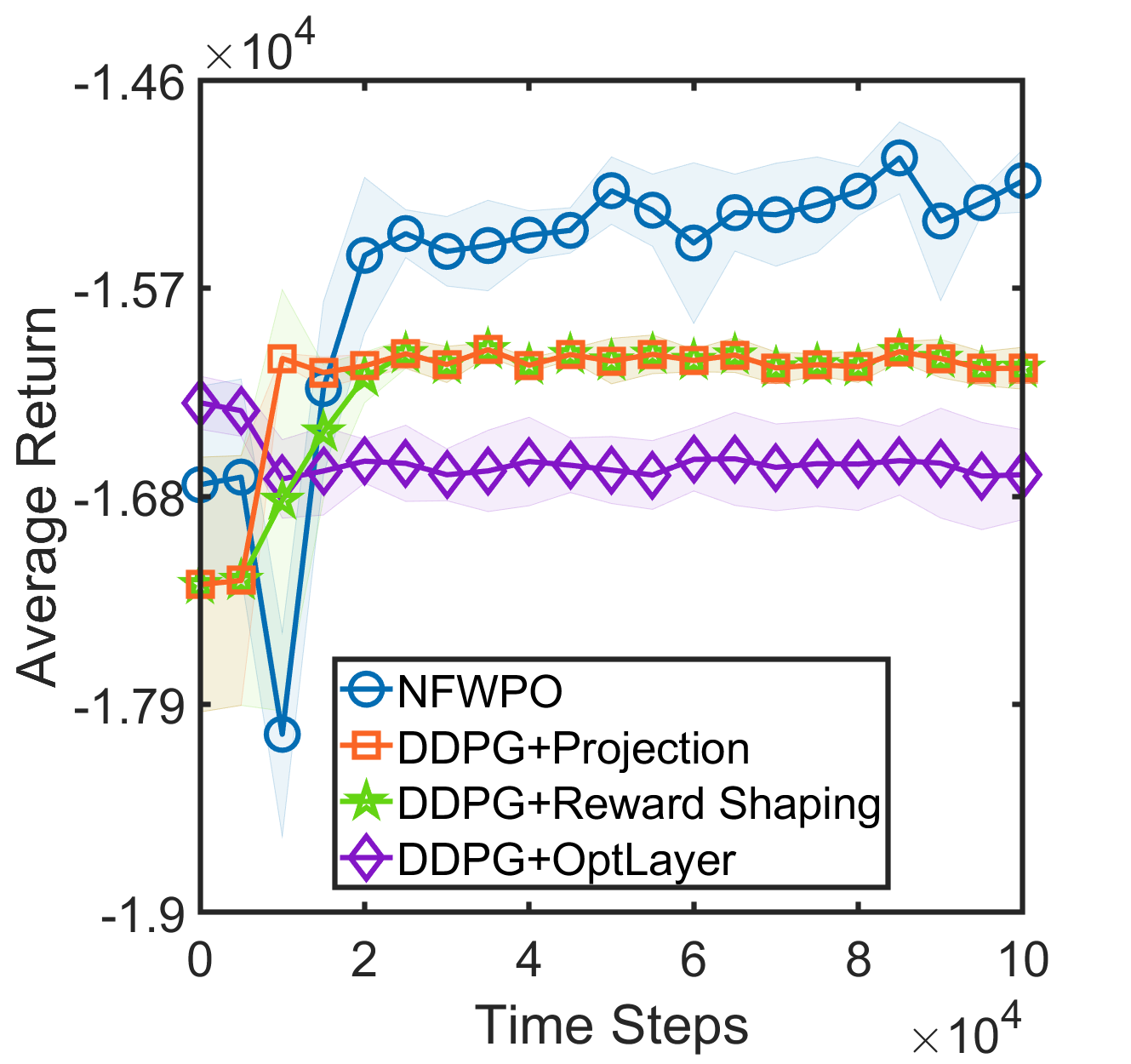}} \label{fig:Bike net reward}
    & \hspace{-5mm} \scalebox{0.255}{\includegraphics[width=\textwidth]{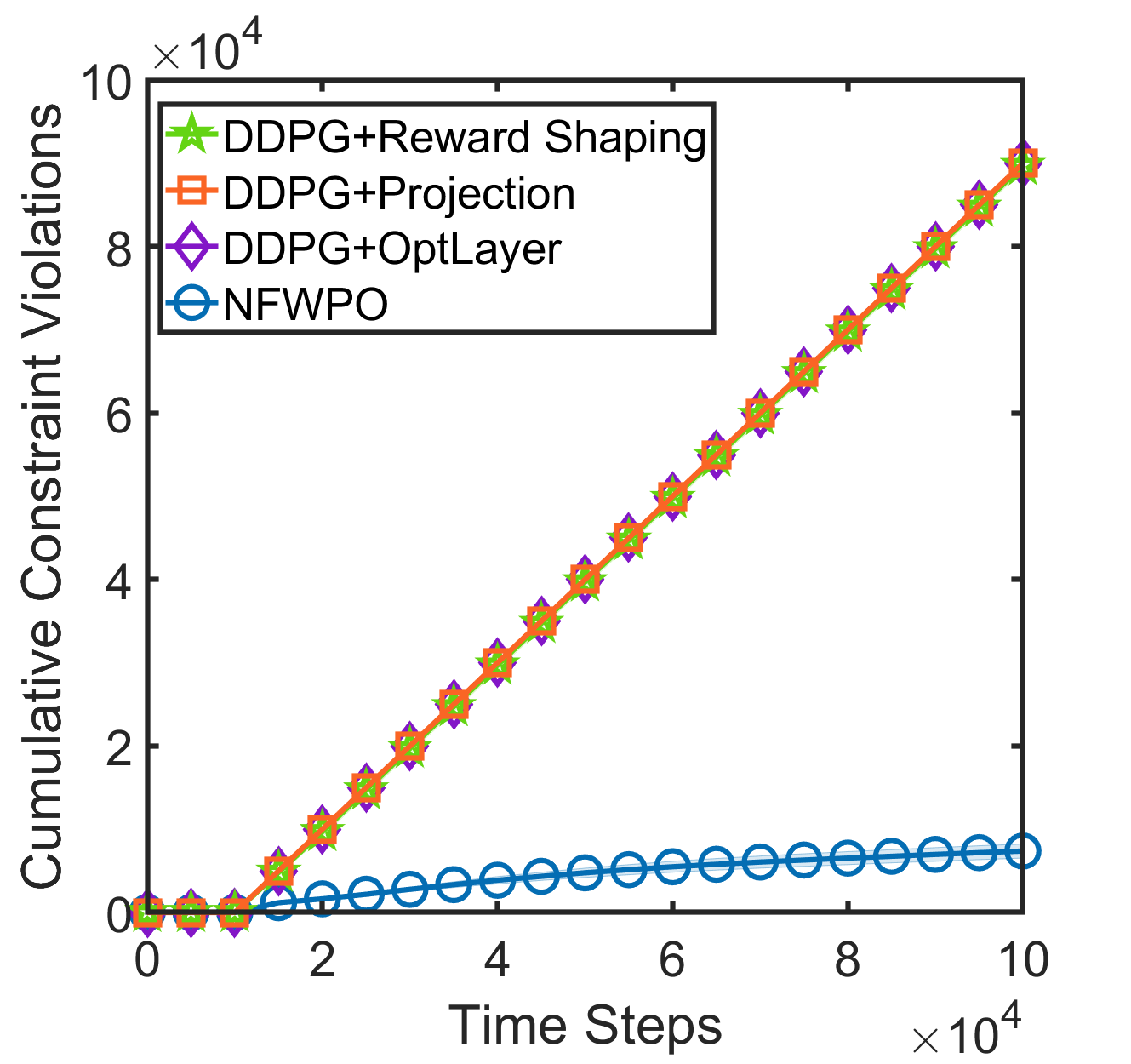}} \label{fig:Bike net cons}
    &\hspace{-5mm} \scalebox{0.255}{\includegraphics[width=\textwidth]{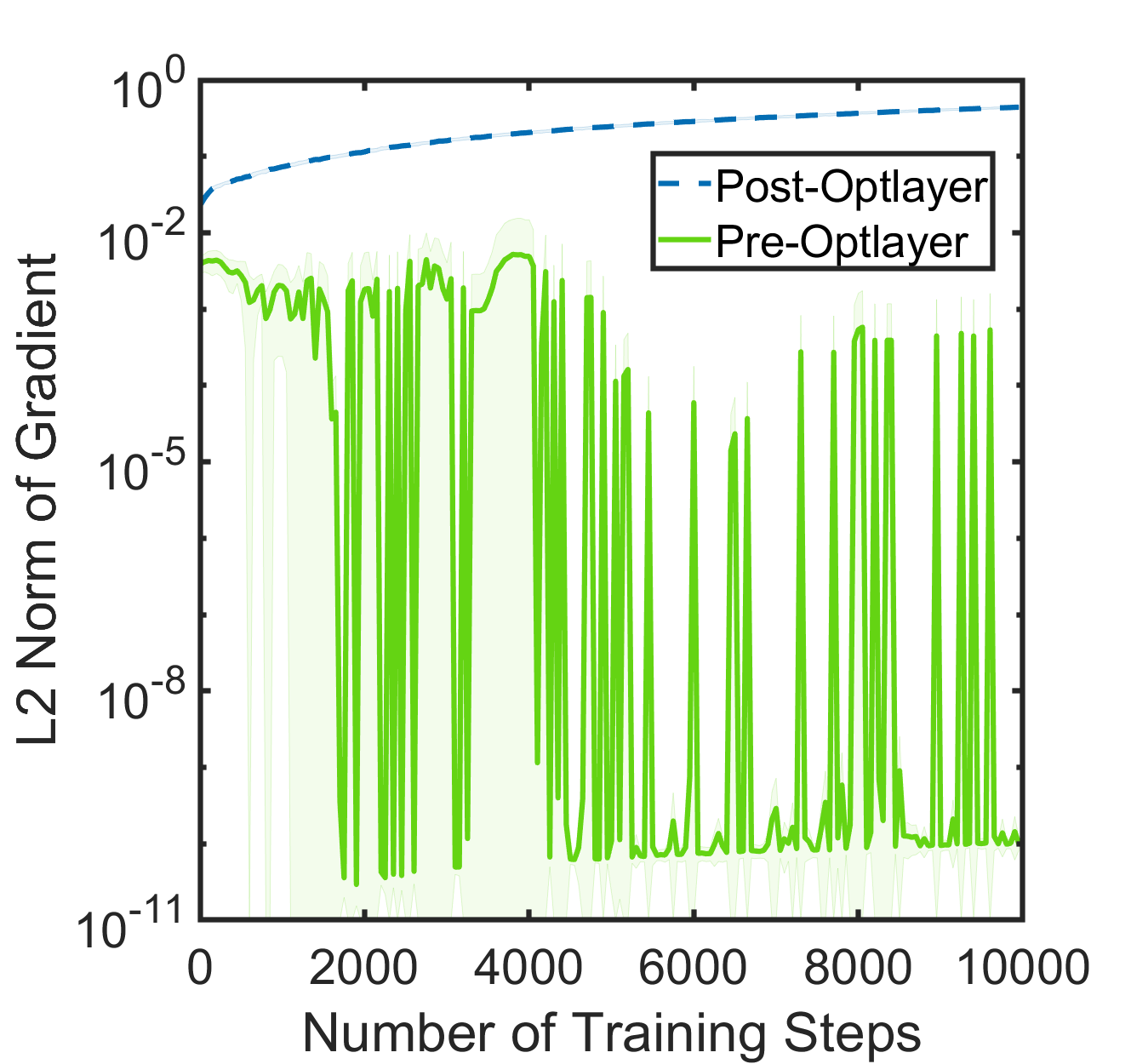}}  \label{fig:Bike Grad}\\
    \mbox{\hspace{-0mm}\small (a)} & \hspace{-2mm} \mbox{\small (b)} & \hspace{-2mm} \mbox{\small (c)} \\
\end{array}$
\caption{Bike sharing problem with $n=5$ (BSS-5) under neural policies: (a) Average return over 5 random seeds; (b) Cumulative number of constraint violations of the pre-projection actions during training; (c) $L_2$-norm of the sample-based gradients with respect to the pre- and post-OptLayer actions of DDPG+OptLayer.}
\label{fig:BSS-5}
\end{figure*}

\vspace{-1mm}
\textbf{Evaluating NFWPO.}
We proceed to compare NFWPO with the other three baselines in solving a larger-scale bike-sharing problem with $m=150$, $n=5$, and $C=35$.
As shown by Figure \ref{fig:BSS-5}(a), NFWPO converges faster and achieves a larger return than the other baselines.
To better understand its behavior, Figure \ref{fig:BSS-5}(b) shows the cumulative constraint violations of the \textit{pre-projection} actions.
Interestingly, the pre-projection actions of NFWPO can largely avoid constraint violation, and thus requires less help from the projection during training.
By contrast, all the baselines rely heavily on the projection step to stay feasible, because most of their pre-projection actions fail to satisfy the constraints.
We also observe that DDPG+Projection and DDPG+RewardShaping attain similar average return and frequency of violation. This is because they both produce pre-projection actions far from the feasible sets and thereby obtain similar post-projection actions.
Meanwhile, DDPG+OptLayer suffers from nearly zero learning progress due to the zero-gradient issue.
Figure \ref{fig:BSS-5}(c) compares the sample-based gradient with respect to the pre and post-OptLayer actions. 
Since the gradients of the pre-OptLayer actions (green line in Figure \ref{fig:BSS-5}(c)) are mostly close to zero, the sample-based policy gradients $\nabla_\theta \hat{J}_{\mu}(\pi(\cdot;\theta))$ are therefore close to zero for most of the training steps.
As the gradients with respect to the post-OptLayer actions are always \textit{non-zero} (blue dotted line in Figure \ref{fig:BSS-5}(c)), we know that the zero-gradient issue of $\nabla_\theta \hat{J}_{\mu}(\pi(\cdot;\theta))$ indeed results from the projection layer.
This confirms that the additional OptLayer could easily lead to the zero-gradient issue and sample-inefficient training.

\vspace{-1mm}
\subsection{Utility Maximization of Communication Networks}
\label{section:exp:network}
\vspace{-1mm}

In this section, we evaluate the proposed methods over the task of utility maximization in communication networks.
We simulate the network with the open-source network simulator from PCC-RL\footnote{PCC-RL: https://github.com/PCCproject/PCC-RL} \citep{jay2019internet}.
For the network topology, we consider the classic T3 NSFNET Backbone and set the bandwidth of each link to be 50 packets per second throughout the experiments.
We generate three network flows, each of which has three candidate paths from its source to the destination. 
The action is to determine the rate allocation of each flow along each candidate path. 
The reward consists of three parts: (i) Throughput: the number of received packets per second; (ii) Drop rate: the number of dropped packets per second; (iii) Latency: the average latency of the packets in the last second. For each flow $i$, its immediate reward is $\log({\frac{\text{throughput}_{\bf i} }{\text{drop rate}_{\bf i}^{0.5} \times \text{latency}_{\bf i}^{1.5}} })$, which corresponds to the widely-used proportional fairness criteria \citep{kelly1997charging}.
One salient feature of a communication network is that when the total packet arrival rate of a link approaches its bandwidth, the latency will grow rapidly, and accordingly most of the packets would be dropped. 
Therefore, in this environment, the action constraints correspond to the link bandwidth constraints, i.e., the total assigned packet arrival rate of each link should be bounded by 50. 

\begin{figure}[!htb]
\centering
$\begin{array}{c c}
    \multicolumn{1}{l}{\mbox{\bf }} & \multicolumn{1}{l}{\mbox{\bf }}   \\ 
    \hspace{-4mm} \scalebox{0.255}{\includegraphics[width=\textwidth]{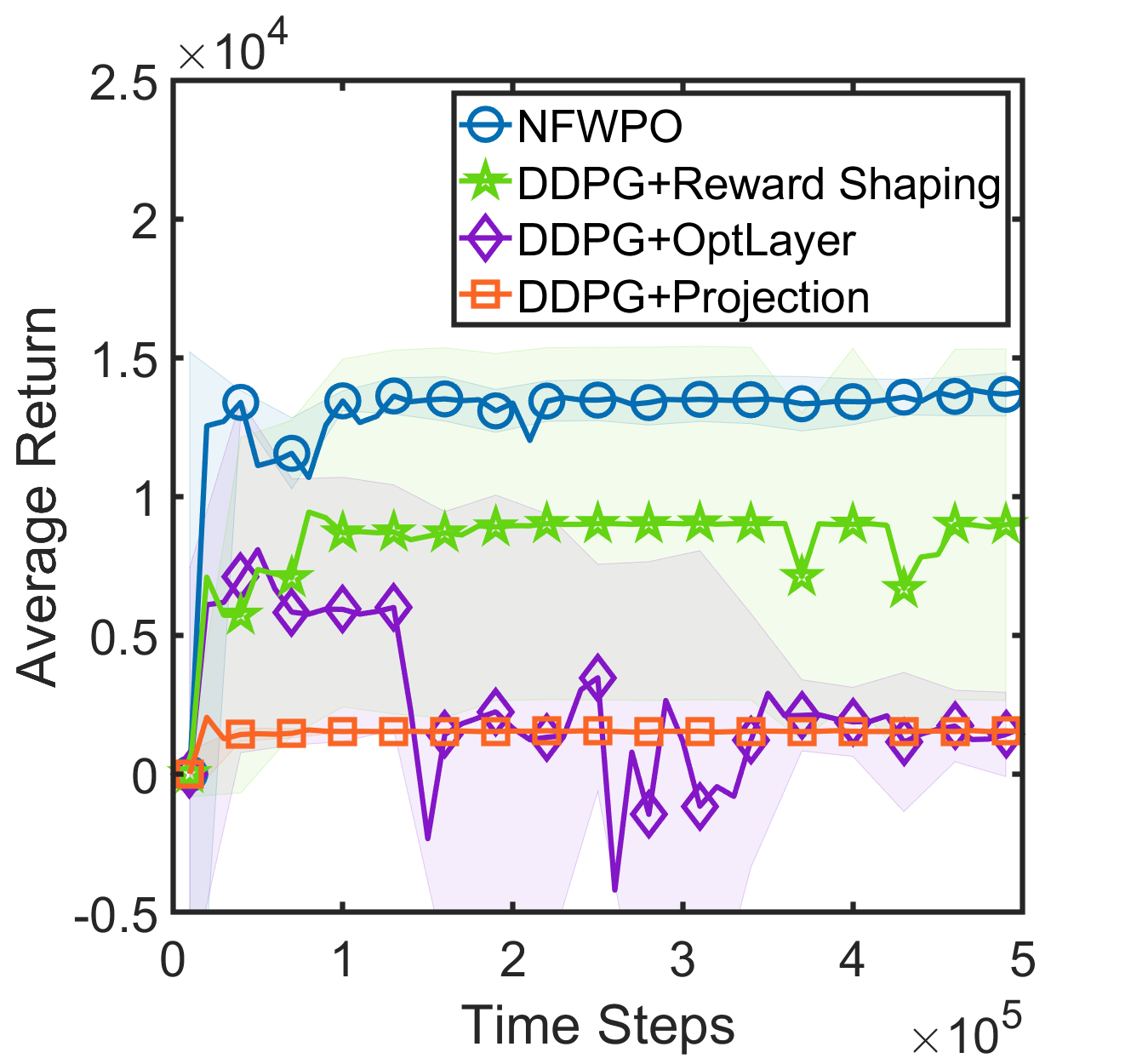}}  \label{fig:Network_reward} & 
    \hspace{-7mm} \scalebox{0.255}{\includegraphics[width=\textwidth]{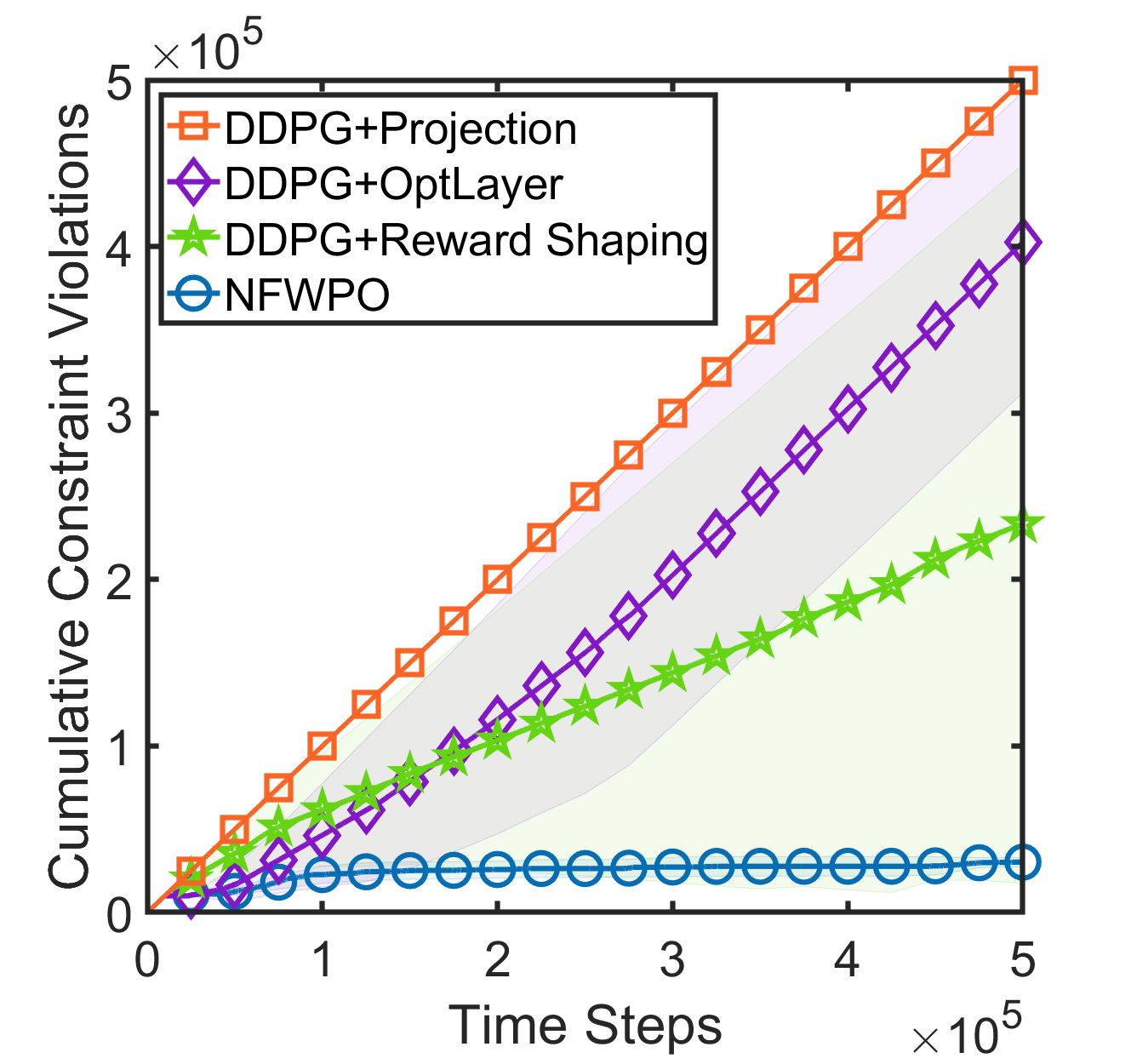}} 
    \label{fig:Network_constraint_violate} \\
    \mbox{\small (a)} & \hspace{-2mm} \mbox{\small(b)}
\end{array}$
\caption{Utility maximization in NSFNET: (a) Average return over 5 random seeds; (b) {Cumulative constraint violations of the pre-projection actions during training.}}
\label{fig:Network all}
\end{figure}

Figure \ref{fig:Network all}(a) shows the training curves and indicates that NFWPO still converges fast (in about $10^5$ steps) and achieves much larger return than the baselines.
Moreover, similar to the bike-sharing problems, we see from Figure \ref{fig:Network all}(b) that most of the pre-projection actions of NFWPO already satisfy the constraints.
In this task, we find that reward shaping does help in guiding the pre-projection actions towards the feasible action sets, but only under some random seeds (and therefore the large variance in Figures \ref{fig:Network all}(a)-(b)).
Regarding DDPG+OptLayer, in the initial training phase, we observe that it mostly produces pre-OptLayer actions with small flow rates, which lead to a smaller number of constraint violations and moderate returns.
To achieve a higher return, DDPG+OptLayer then gradually increases the flow rates but accidentally causes more constraint violations of pre-OptLayer actions and suffers from the inaccurate gradient issue described in Section \ref{section:exp:BSS}.
Ultimately, DDPG+OptLayer can only achieve a fairly low return.
 
\vspace{-2mm}

\begin{table*}[!htb]
    \centering
    \caption{Average return over the final 10 evaluations.}
    \label{tab:expResult}
    \begin{tabular}{l c c c}
         \hline Methods & BSS-3 & BSS-5 & NSFNET \\
         \hline
         NFWPO & \textbf{-1673.04} & \textbf{-15132.21} & \textbf{13770.67} \\
         DDPG+Projection & -2254.52 & -16123.48 & 1514.44\\
         DDPG+Reward Shaping & -2308.00 & -16123.48 & 9010.46 \\
         DDPG+OptLayer & - & -16686.04 & 1667.59 \\
         \hline
    \end{tabular}
\end{table*}

\begin{table*}[!htb]
    \centering
    \caption{Average return over the final 10 evaluations in the MuJoCo environments.}
    \label{tab:mujocoResult}
    \begin{tabular}{l c c}
         \hline Methods & Reacher & Halfcheetah \\
         \hline
         NFWPO & \textbf{-4.76} & \textbf{6513.26} \\
         DDPG+Projection & -11.15 & 2746.72 \\
         DDPG+Reward Shaping &  -8.66 & 3065.37 \\
         DDPG+OptLayer & -7.25 & 1399.37 \\
         SAC+Projection & -10.50 & 4874.45 \\
         TRPO+Projection & -11.04 & 2247.82 \\          PPO+Projection & -10.68 &  1459.04 \\
         FOCOPS+Projection & -12.23 & 1916.46 \\
      \hline
    \end{tabular}
\end{table*}

\vspace{-1mm}
\subsection{MuJoCo Continuous Control Tasks}
\label{section:exp:Mujoco}
\vspace{-1mm}
To further validate NFWPO, we consider popular continuous control tasks in MuJoCo \citep{6386109} with non-linear and state-dependent action constraints.
We further compare the proposed algorithm with important benchmarks RL algorithms for MuJoCo control taks, including PPO \citep{schulman2017proximal}, TRPO \citep{schulman2015trust}, and SAC \citep{haarnoja2018soft}.
To make the comparison even more comprehensive, we also evaluate FOCOPS \citep{zhang2020first}, which is a recent approach designed to address long-term total discounted cost constraints. 
As FOCOPS is not designed for handling state-wise action constraints, we relax the action constraints into the long-term total discounted constraints required by FOCOPS.
To ensure constraint satisfaction, we use the same technique as DDPG+Projection, i.e., the actions of PPO, TRPO, SAC, and FOCOPS are post-processed by $L_2$-projection before being applied to the environment.

\begin{figure}[!htb]
\centering
$\begin{array}{c c}
    \multicolumn{1}{l}{\mbox{\bf }} & \multicolumn{1}{l}{\mbox{\bf }}   \\ 
    \hspace{-4mm} \scalebox{0.255}{\includegraphics[width=\textwidth]{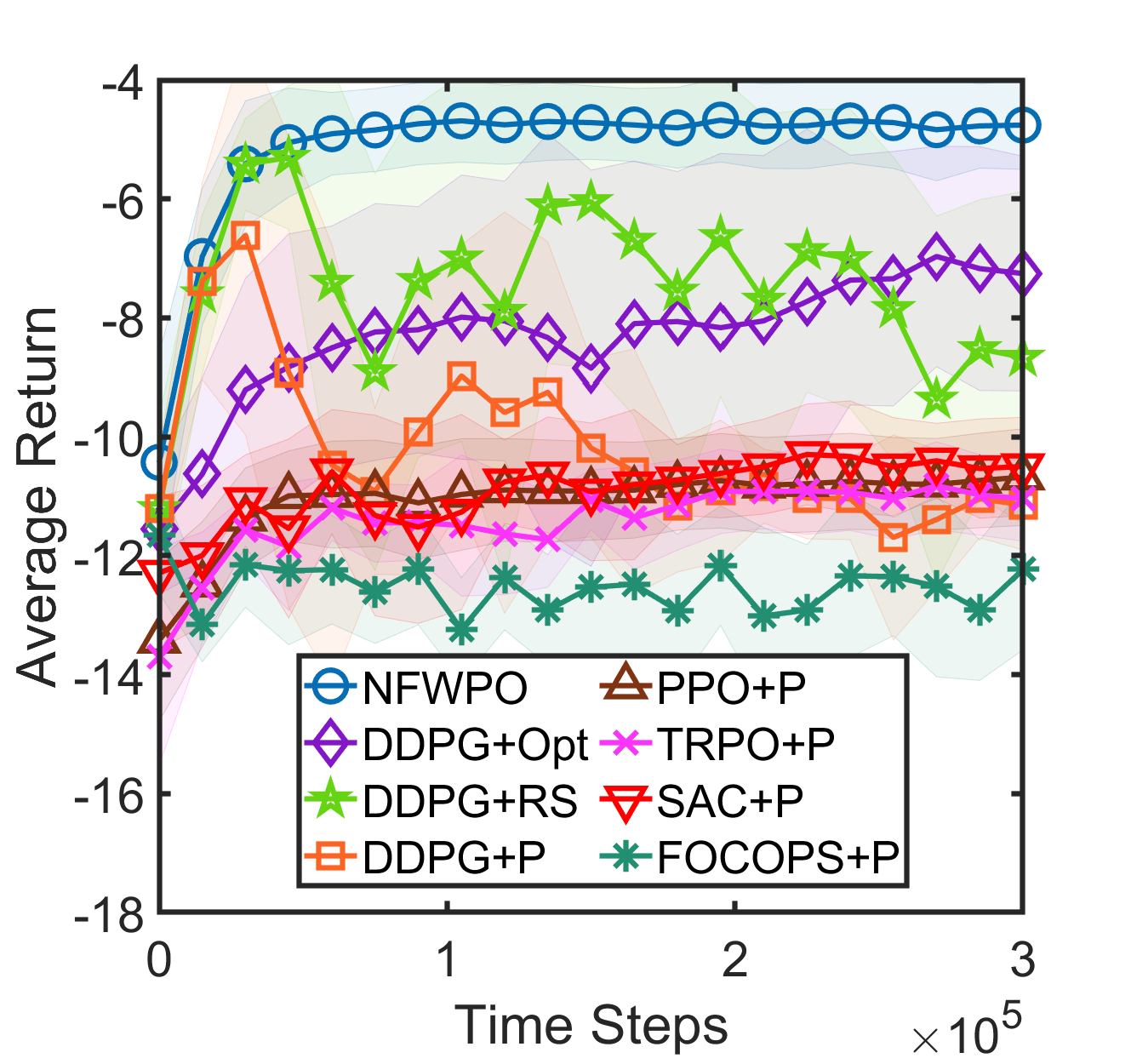}}  \label{fig:Reacher_reward} & 
    \hspace{-6mm} \scalebox{0.255}{\includegraphics[width=\textwidth]{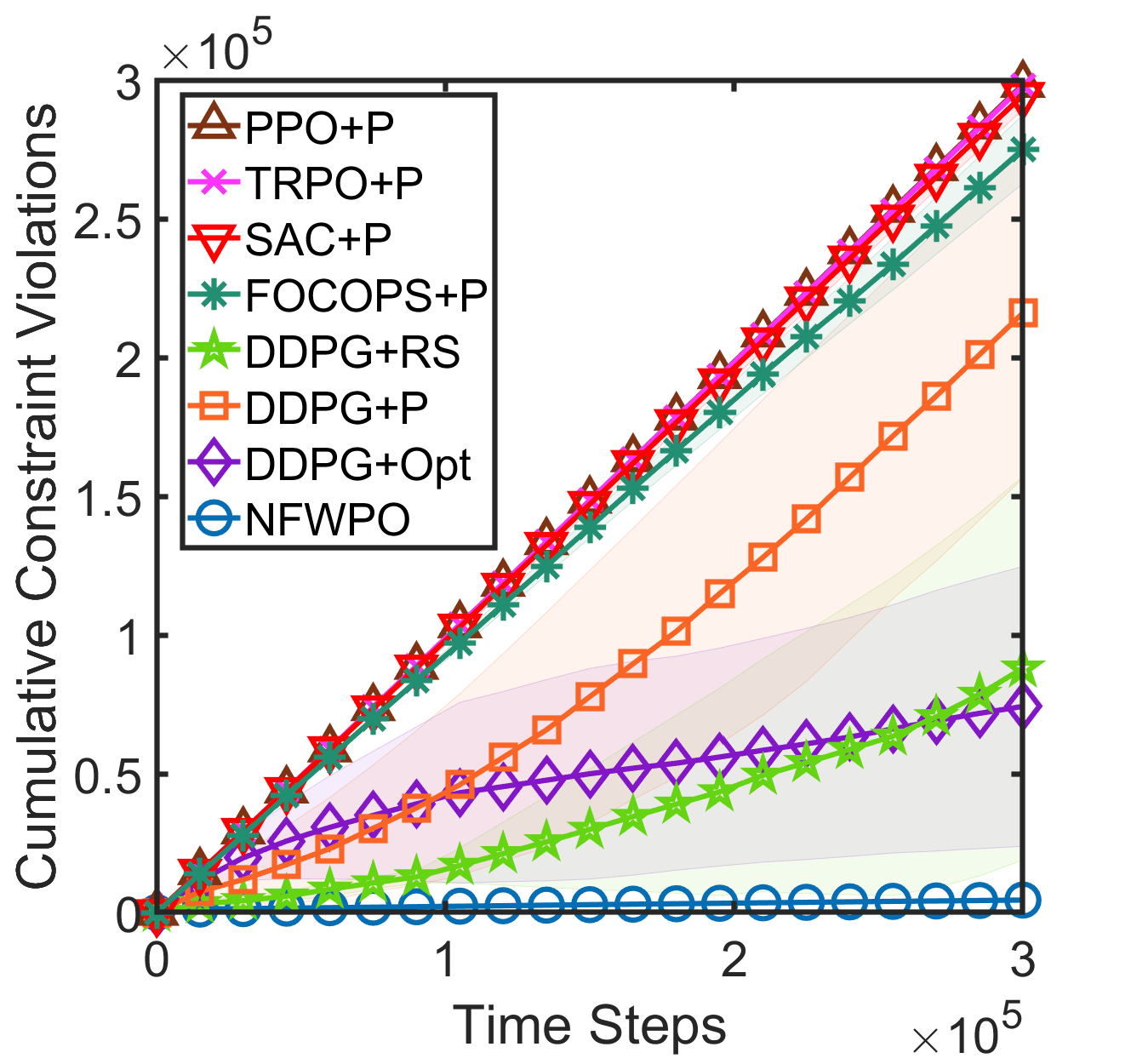}}
    \label{fig:Reacher_constraint_violate} \\
    \mbox{\small (a)} & \hspace{-2mm} \mbox{\small(b)}
\end{array}$
\caption{Reacher-v2 with a non-linear action constraint: (a) Average return over 5 random seeds; (b) {Cumulative constraint violations of the pre-projection actions during training (``+P'' stands for ``+Projection'', ``+RS'' stands for ``Reward Shaping'', and ``+Opt'' stands for ``+OptLayer'').}}
\label{fig:Reacher all}
\end{figure}

\vspace{-1mm}
\noindent \textbf{Reacher with non-linear constraints.} In this task, the action space is $2$-dimensional (denoted by $u_{1},u_{2}$), and each action entry corresponds to the torque of a joint of a 2-DoF robot.
To validate the applicability of NFWPO, we impose nonlinear action constraints as: $ {u_{1}}^2+{u_{2}}^2\leq 0.05$.
From Figure \ref{fig:Reacher all}(a), we observe a similar trend that NFWPO still converges faster and achieves a larger return than the other baseline methods.
In this task, DDPG+OptLayer and DDPG+RewardShaping can achieve a return closer to NFWPO as it has fewer pre-OptLayer (pre-projection) violations as shown in Figure \ref{fig:Reacher all}(b).
The other algorithms still perform poorly as they always produce actions far from the feasible sets and relies heavily on the projection step. 
This is reflected by the fact that SAC, TRPO, and PPO violate the constraint at almost every training step, as shown in Figure \ref{fig:Reacher all}(b). Regarding FOCOPS, the violation of the constraint in the second half of training is less than that in the first half.
This indicates that FOCOPS needs much more training steps to find a policy that violate lesser,
Despite this, the return of FOCOPS remains fairly low as other benchmarks.

\vspace{-1mm}

\begin{figure}[!tb]
\centering
$\begin{array}{c c}
    \multicolumn{1}{l}{\mbox{\bf }} & \multicolumn{1}{l}{\mbox{\bf }}   \\ 
    \hspace{-4mm} \scalebox{0.258}{\includegraphics[width=\textwidth]{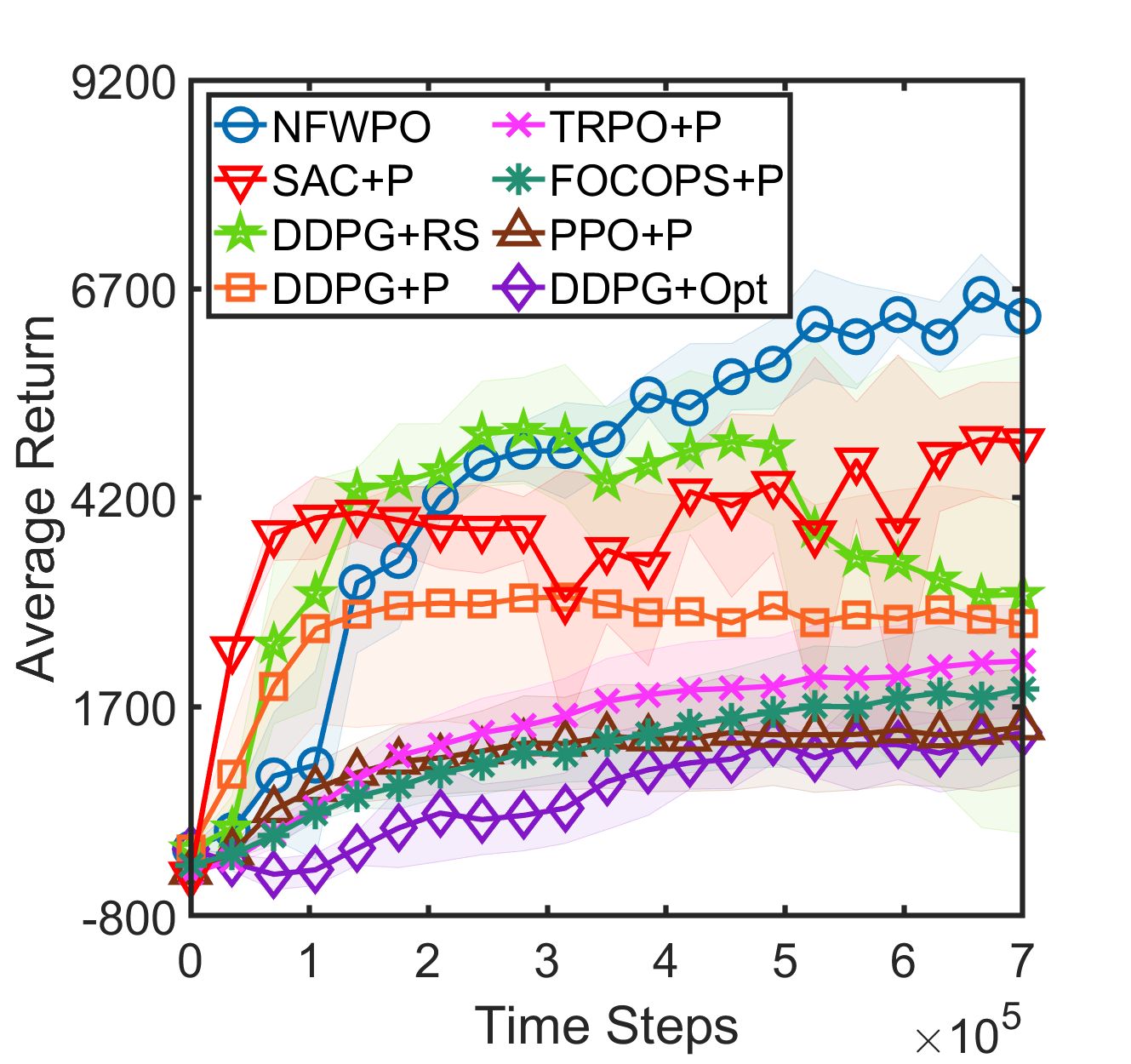}}  \label{fig:Halfcheetah_reward} & 
    \hspace{-6mm} \scalebox{0.258}{\includegraphics[width=\textwidth]{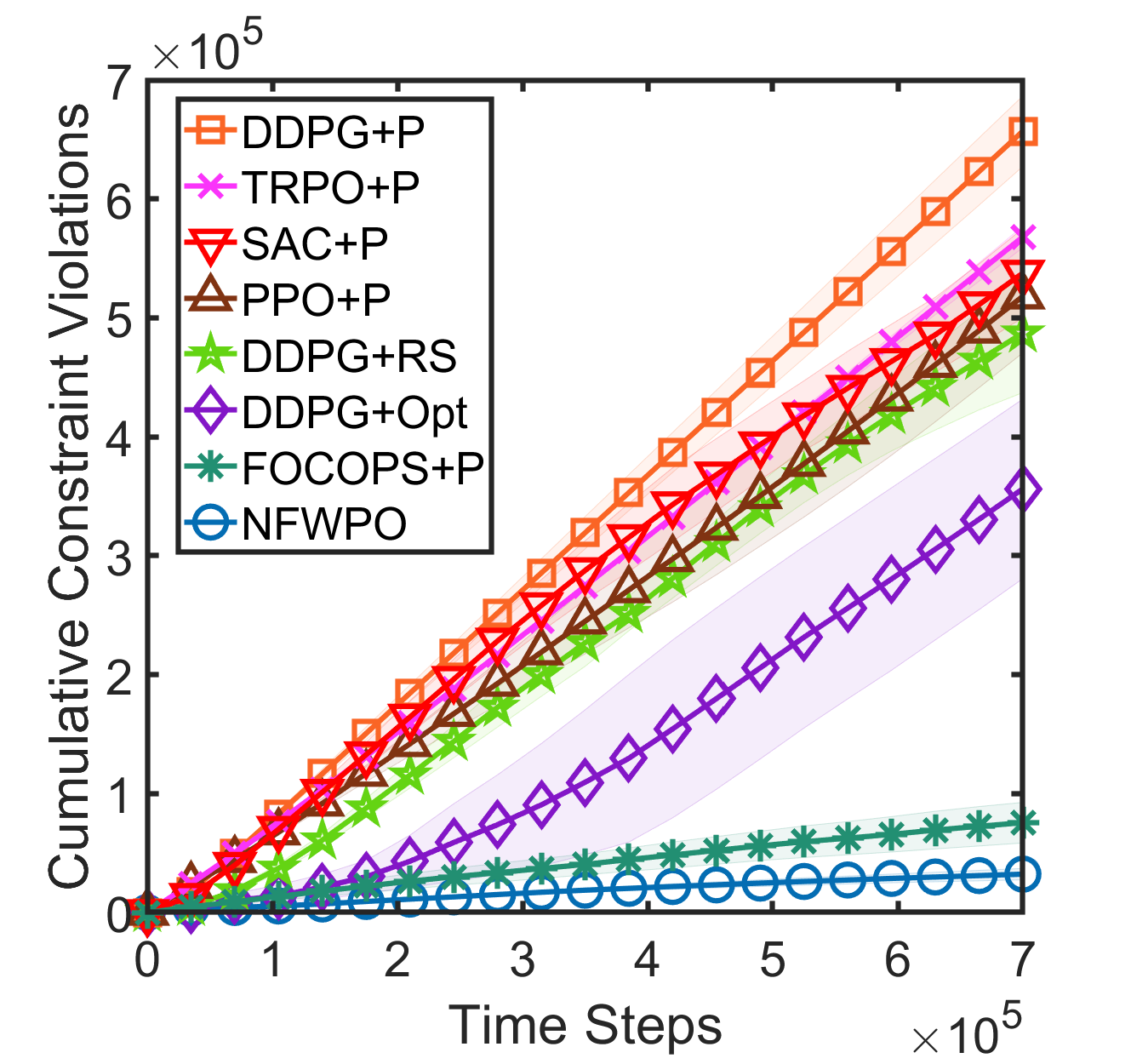}}
    \label{fig:Halfcheetah_constraint_violate} \\
    \mbox{\small (a)} & \hspace{-2mm} \mbox{\small(b)}
\end{array}$
\caption{Halfcheetah-v2 with a state-dependent action constraint: (a) Average return over 5 random seeds; (b) {Cumulative constraint violations of the pre-projection actions during training (``+P'' stands for ``+Projection'', ``+RS'' stands for ``Reward Shaping'', and ``+Opt'' stands for ``+OptLayer'').}}
\label{fig:Halfcheetah all}
\end{figure}

\noindent \textbf{Halfcheetah with state-dependent constraints.} In this task, an action is $6$-dimensional and is denoted by $(v_{1},\cdots,v_{6})$.
We consider a challenging scenario where the constraint is state-dependent. Specifically, the imposed constraint is $\sum_{i=1}^{6} |v_i w_i|\leq 20$, where $w_i$ denotes the angular velocity of the $i$-th joint and is part of the state. This constraint is meant to capture the limitation of total output power.
Similar to the other environments, from Figure \ref{fig:Halfcheetah all}(a)-(b), we still observe that NFWPO achieves better sample efficiency than the other baselines. Moreover, NFWPO violates the constraint for only 3\% of the steps while getting the highest return. 
On the other hand, PPO, TRPO, and SAC all violate the constraint for more than 75\% of the steps. FOCOPS violates the constraint for about 15\% of the time but only achieves a fairly low return. 
Again, we see that NFWPO outperforms all the baseline methods with much less constraint violation.

\vspace{-1mm}

%% file: 02-related.tex
\vspace{-2mm}
\section{Related Work}
\label{section:related}
\vspace{-2mm}

The constrained RL problems have been extensively studied from two main perspectives.
The first category encodes the constraints via cost signals, which are incurred at each step along with the reward signals, and accordingly focuses on the average-cost constraints.
This line of research works borrows a variety of ideas from the optimization literature.
For example, \citep{chow2015risk} addressed chance constraints by using a primal-dual approach to achieve a trade-off between return and risk.
Similarly, \citep{tessler2018reward} proposed Reward Constrained Policy Optimization which applied Lagrangian relaxation and converted the constraints into penalty terms in the objective.
\citep{achiam2017constrained} proposed Constrained Policy Optimization to achieve strict policy improvement under the average cost constraints by using the trust-region approach. 
In \citep{chow2018lyapunov}, Lyapunov-based safe reinforcement learning was proposed to address the constraints by solving a linear program.
\citep{yang2019projection} proposed Projection-Based Constrained Policy Optimization to achieve no constraint violation by taking a projection step after the local reward improvement update.
In \citep{liu2020ipo}, Interior-point Policy Optimization was proposed to handle the average cost constraints by augmenting the objective with logarithmic barrier functions.
\citep{satija2020constrained} took a different approach by converting the trajectory-level constraints into per-step state-wise constraints and accordingly defining a safe policy improvement step.
\citep{zhang2020first} proposed an approach to solving the constrained policy optimization problem by first finding a target policy directly in the policy space and thereafter converting the target policy to a parameterized one through a projection step onto the parameter space.
Different from all the above prior works, this paper considers the RL settings with state-wise action constraints.

The second category is on the state-wise constraints that need to be satisfied on a step-by-step basis. 
\citep{pham2018optlayer} studied the state-wise action constraints of robotic systems and proposed a projection-based OptLayer to enforce the constraints.
\citep{dalal2018safe} also considered state-wise safety constraints under linearization and proposed a projection-based safety layer to handle the constraints.
Similarly, \citep{bhatia2019resource} considered resource constraints and proposed variants of OptLayer to improve the computational efficiency.
\citep{shah2020solving} proposed a more efficient projection scheme for enforcing linear action constraints.
Despite the similarity in the problem setting, we take a different approach and propose a decoupling framework by leveraging Frank-Wolfe to address the action constraints and completely avoid the zero-gradient issue.



%% file: 08-conclusion.tex
\vspace{-2mm}
\section{Conclusion}
\label{section:conclusion}
\vspace{-2mm}

This paper revisits action-constrained RL to tackle the zero-gradient issue and ensure zero constraint violation simultaneously.
We achieve this goal by developing a learning framework that decouples the policy parameter update from constraint satisfaction by leveraging state-wise Frank-Wolfe and a regression argument.
Our theoretical and experimental results demonstrate that the proposed learning algorithm is indeed a promising approach for action-constrained RL.

%% file: appendix.tex
\newpage
\appendix
\onecolumn
\setcounter{section}{0}
\section{Proofs of the Theoretical Results}
\label{section:appendix:proof}
In this section, we provide the proof of the convergence result of the tabular FWPO in Proposition \ref{PROP:CONVERGENCE} by leveraging the proof technique of smooth optimization. 
We start by establishing the key smoothness result of Proposition \ref{PROP: L-SMOOTH OF J} and then proceed to the convergence analysis.

\subsection{Smoothness Results and Proof of Proposition \ref{PROP: L-SMOOTH OF J}}
\label{section:appendix:smoothness}

In this section, we show the smoothness results of the value functions as well as the performance objective under the regularity assumptions (A1)-(A2).
Notably, given the regularity conditions of $r$ and $p$ in action, it remains non-trivial to establish the smoothness of $V(\cdot;\theta)$ and $J_{\mu}(\pi(\cdot;\theta))$ in $\theta$ due to the multi-step compound effect of the changes in policy parameters on the value functions. 
Despite this challenge, we are still able to characterize the required smoothness conditions.

Recall from Definition \ref{def:L smoothness} that a differentiable function $f:\text{dom}f \rightarrow \bR$ is $L_0$-smooth if there exists $L_0\geq 0$ such that for any $x,y\in \text{dom}f$, we have $\norm{\nabla f(x) - \nabla f(y)}_2\leq L_0\norm{x-y}_2$.
One useful property is that if $f$ is $L_0$-smooth, then $f$ satisfies
\begin{equation}
    \lvert f(y)-f(x)- \inp{\nabla f(x)}{y-x}\rvert  \leq \frac{L_0}{2}\norm{y-x}_2^2, \hspace{6pt} \forall x,y\in \text{dom}f.\label{eq:L-smooth bound}
\end{equation}

For notational convenience, we explicitly number the state space as $\cS=\{1,\cdots,M\}$. 
By slightly abusing the notation, we use $\pi$ to denote the $M\times N$ matrix where the $s$-th row represents the action selected by a deterministic policy $\pi$ at each state $s\in\{1,\cdots,M\}$.
Let $\bP^{\pi}$ denote a $M \times M$ matrix where the $(i,j)$-th entry is $p(j\rvert i,\pi(i))$.
Given a deterministic policy $\pi$, consider a ``perturbed'' version of the policy defined by $\pi_{\delta}=\pi+\delta \bW$, where $\delta\in\bR$ and $\bW\in\bR^{M\times N}$ is some fixed matrix with $\norm{\bW}_F=1$.
Moreover, we simplify the notations by letting $\bP(\delta)\equiv\bP^{\pi_\delta}$ and $V_{s}(\delta)\equiv V(s;\pi_{\delta})$ and then define $\bG(\delta):=(\bI_{M}-\gamma \bP(\delta))^{-1}$, where $\bI_{M}$ denotes the identity matrix of size $M\times M$.
To show the smoothness result of $V(s;\theta)$, it is sufficient to show that $\lvert\frac{d^2 V_s(\delta)}{d\delta^2}\rvert$ is bounded for any $\bW\in\bR^{M\times N}$ with $\norm{\bW}_F=1$, for any state $s$.
We first introduce several useful lemmas.

\begin{lemma}[Matrix-by-scalar derivatives]
\label{lemma:matrix derivative}
\begin{align}
    \frac{d \bG(\delta)}{d\delta}&=\gamma\bG(\delta)\frac{d \bP(\delta)}{d \delta}\bG(\delta),\label{eq:matrix derivative 1}\\
    \frac{d^2 \bG(\delta)}{d\delta^2}&=2\gamma\bG(\delta)\frac{d \bP(\delta)}{d \delta}\bG(\delta)\frac{d \bP(\delta)}{d \delta}\bG(\delta)+ \bG(\delta)\frac{d^2 \bP(\delta)}{d \delta^2}\bG(\delta).\label{eq:matrix derivative 2}
\end{align}

\end{lemma}
\begin{myproof}[Lemma \ref{lemma:matrix derivative}]
\normalfont By the definition of $\bG(\delta)$, we know
\begin{equation}
    \bI_M=\bG(\delta)(\bI_M-\gamma \bP(\delta))=\bG(\delta)-\gamma \bG(\delta) \bP(\delta).\label{eq:I=G times Ginv} 
\end{equation}
Note that the product rule of matrix-by-scalar derivative suggests that for two matrices $\bU_1(\delta),\bU_2(\delta)$, we have $\frac{d}{d\delta}(\bU_1\bU_2)=\bU_1\frac{d \bU_2}{d\delta}+\frac{d \bU_1}{d\delta}\bU_2$. By taking the matrix-by-scalar derivative with respect to $\delta$ on both sides of (\ref{eq:I=G times Ginv}) and using the product rule, we have
\begin{equation}
  \frac{d \bG(\delta)}{d \delta}-\Big(\gamma \bG(\delta)\frac{d \bP(\delta)}{d \delta}+\gamma\frac{d \bG(\delta)}{d \delta}\bP(\delta)\Big)=\b0_M,\label{eq:matrix derivative 3}
\end{equation}
where $\b0_M$ denotes the $M\times M$ zero matrix.
By reorganizing the terms in (\ref{eq:matrix derivative 3}), it is straightforward to verify that (\ref{eq:matrix derivative 1}) holds.
Based on (\ref{eq:matrix derivative 1}), we can obtain the second-order derivative of $\bG(\delta)$ as
\begin{align}
     \frac{d^2 \bG(\delta)}{d\delta^2}&=\frac{d}{d\delta}\Big(\bG(\delta)\frac{d \bP(\delta)}{d \delta}\bG(\delta)\Big)\\
     &=\bG(\delta)\frac{d}{d\delta}\Big(\frac{d \bP(\delta)}{d \delta}\bG(\delta)\Big)+ \frac{d \bG(\delta)}{d\delta}\Big(\frac{d \bP(\delta)}{d \delta}\bG(\delta)\Big)\\
     &=2\gamma\bG(\delta)\frac{d \bP(\delta)}{d \delta}\bG(\delta)\frac{d \bP(\delta)}{d \delta}\bG(\delta)+ \bG(\delta)\frac{d^2 \bP(\delta)}{d \delta^2}\bG(\delta).
\end{align}
Hence, we conclude that (\ref{eq:matrix derivative 1}) and (\ref{eq:matrix derivative 2}) indeed hold. \MYQED
\end{myproof}

\begin{lemma}
\label{lemma:infinite norm of G(z)x}
For any $x\in\bR^{M}$, $\bG(\delta)$ satisfies that
\begin{equation}
    \norm{\bG(\delta)x}_{\infty}\leq \frac{1}{1-\gamma}\norm{x}_{\infty}.
\end{equation}
\end{lemma}
\begin{myproof}[Lemma \ref{lemma:infinite norm of G(z)x}]
\normalfont Note that $\bG(\delta)=(\bI_{n}-\gamma \bP(\delta))^{-1}=\sum_{m=0}^{\infty}\gamma^m \bP(z)^{m}$.
Given that $\bP(\delta)$ is a stochastic matrix where all the elements are non-negative and each row sums to 1, we know $\bP(\delta)^m$ is also a stochastic matrix, for any $m\in \mathbb{N}$.
For each $i$, let $e_i$ denote an $n$-dimensional one-hot vector with the $i$-th entry equal to $1$. 
Then, it is easy to verify that $\norm{\bG(\delta)e_i}_{\infty}\leq\frac{1}{1-\gamma}$, for any $i\in \{1,\cdots,M\}$.
This thereby implies that $\norm{\bG(\delta)x}_{\infty}\leq \frac{1}{1-\gamma}\norm{x}_{\infty}$, for any $x\in\bR^M$.\MYQED
\end{myproof}

\begin{lemma}
\label{lemma:norm bound of derivative of P}
Under the regularity assumptions (A1)-(A2), there exist constants $C_1,C_2>0$ such that for any $\bW\in\bR^{M\times N}$ with $\norm{\bW}_F=1$ and for any $x\in \bR^{M}$, we have
\begin{align}
    \norm{\frac{d \bP(\delta)}{d\delta}x}_{\infty}&< C_1 \norm{x}_{\infty},\label{eq:norm bound of derivative of P}\\
    \norm{\frac{d^2 \bP(\delta)}{d\delta^2}x}_{\infty}&< C_2 \norm{x}_{\infty}.\label{eq:norm bound of 2nd derivative of P}
\end{align}
\end{lemma}
\begin{myproof}[Lemma \ref{lemma:norm bound of derivative of P}]
\normalfont For convenience, we use $\bW_i$ to denote the $i$-th row of $\bW$. For any pair of $i,j\in \{1,\cdots,M\}$, the $(i,j)$-th element of $\frac{d \bP(\delta)}{d \delta}$ evaluated at $\delta=0$ satisfies
\begin{align}
    \bigg\lvert\Big[\frac{d \bP(\delta)}{d \delta}\Big\rvert_{\delta=0}\Big]_{(i,j)}\bigg\rvert&=\bigg\lvert\lim_{h\rightarrow 0}\frac{p(j\rvert i, \pi_h(i))-p(j\rvert i,\pi(i))}{h}\bigg\rvert\label{eq:norm bound of derivative of P 1}\\
    &=\bigg\lvert\inp{\nabla_{a}p(j\rvert i,a)\big\rvert_{a=\pi(i)}}{\bW_i^{\top}}\bigg\rvert\label{eq:norm bound of derivative of P 2}\\
    &\leq \norm{\nabla_{a}p(j\rvert i,a)\big\rvert_{a=\pi(i)}}_2 \cdot\norm{\bW_i}_2\label{eq:norm bound of derivative of P 3}\\
    &< C_p,\label{eq:norm bound of derivative of P 4}
\end{align}
where (\ref{eq:norm bound of derivative of P 2}) follows from the differentiability of $p$ and the property of directional derivatives, (\ref{eq:norm bound of derivative of P 3}) holds by the Cauchy-Schwarz inequality, and (\ref{eq:norm bound of derivative of P 4}) follows from the regularity assumptions and that $\norm{\bW}_F=1$.
Then, by (\ref{eq:norm bound of derivative of P 4}), it is straightforward to verify that (\ref{eq:norm bound of derivative of P}) indeed holds. 

Regarding (\ref{eq:norm bound of 2nd derivative of P}), we first let $\bH_{a}(i,j)$ denote the Hessian of $p(j\rvert i,a)$ with respect to $a$.
The second directional derivative $\frac{d^2 \bP(\delta)}{d \delta^2}$ satisfies that
\begin{align}
    \bigg\lvert\Big[\frac{d^2 \bP(\delta)}{d \delta^2}\Big\rvert_{\delta=0}\Big]_{(i,j)}\bigg\rvert&=\Big\lvert \bW_i \bH_{a}(i,j)\big\rvert_{a=\pi(i)} \bW_i^{\top}\Big\rvert\label{eq:norm bound of 2nd derivative of P 1}\\
    &\leq \norm{\bW_i}_2\cdot \norm{\bH_{a}(i,j)\big\rvert_{a=\pi(i)} \bW_i^{\top}}_2\label{eq:norm bound of 2nd derivative of P 2}\\
    &\leq \norm{\bH_{a}(i,j)\big\rvert_{a=\pi(i)}}_{F} \label{eq:norm bound of 2nd derivative of P 3}\\
    &< L_p M^2,\label{eq:norm bound of 2nd derivative of P 4}
\end{align}
where (\ref{eq:norm bound of 2nd derivative of P 1}) holds by the basic property of second directional derivatives, (\ref{eq:norm bound of 2nd derivative of P 2}) is due to the Cauchy-Schwarz inequality, (\ref{eq:norm bound of 2nd derivative of P 3}) follows from that $\norm{\bW}_F\leq 1$ and $\norm{\bH_{a}(i,j)\rvert_{a=\pi(i)}}_2\leq \norm{\bH_{a}(i,j)\rvert_{a=\pi(i)}}_F$, and (\ref{eq:norm bound of 2nd derivative of P 4}) holds by the $L_p$-smoothness of $p$.
Therefore, by (\ref{eq:norm bound of 2nd derivative of P 4}) we conclude that (\ref{eq:norm bound of 2nd derivative of P}) holds.
\MYQED
\end{myproof}

Now we are ready to establish the smoothness conditions of the value functions.
Define $L_{Q}=L_r+M L_p\frac{\gamma}{1-\gamma}$.
\begin{lemma}
\label{lemma:Q and V L-smooth}
Under the regularity assumptions (A1)-(A2) and tabular direct policy parameterization, we have the following smoothness properties of $Q(s,a;\pi)$ and $V(s;\pi)$:

\noindent (i) Under any fixed policy $\pi$, $Q(s,a;\pi)$ is $L_{Q}$-smooth in action $a$, for any state $s\in \cS$. 

\noindent (ii) There exists some constant $L>0$ such that $V(s;\pi_\theta)$ is $L$-smooth in the policy parameters $\theta$, for any state $s\in\cS$.

\end{lemma}
\begin{myproof}[Lemma \ref{lemma:Q and V L-smooth}]
\normalfont 

\underline{\textbf{For (i):}} Recall that the action-value function can be expressed as
\begin{equation}
    Q(s,a;\pi)=r(s,a)+\gamma \sum_{s'\in\cS}p(s'\rvert s,a)V(s';\pi).\label{eq:Q as r+V}
\end{equation}
By taking the derivative of (\ref{eq:Q as r+V}) with respect to $a$, we have
\begin{equation}
    \nabla_{a} Q(s,a;\pi)=\nabla_a r(s,a)+\gamma \sum_{s'\in\cS}\nabla_{a}p(s'\rvert s,a)V(s';\pi).\label{eq:grad of Q as grad of r + grad of V}
\end{equation}
By the regularity assumptions of $r$ and $p$, we know that for any state $s$ and any two actions $a'$ and $a''$,
\begin{align}
    &\norm{\nabla_{a}Q(s,a;\pi)\rvert_{a=a'} - \nabla_{a}Q(s,a;\pi)\rvert_{a=a''}}_2\label{eq:grad of Q 1}\\
    &\leq \norm{\nabla_{a} r(s,a)\rvert_{a=a'}-\nabla_{a} r(s,a)\rvert_{a=a''}}_2+\gamma \sum_{s'\in\cS}\Big(\norm{\nabla_{a}p(s'\rvert s,a)\rvert_{a=a'}-\nabla_{a}p(s'\rvert s,a)\rvert_{a=a''}}_2\cdot\lvert V(s';\pi)\rvert\Big)\label{eq:grad of Q 2}\\
    &\leq L_r \norm{a'-a''}_2+M L_p \frac{\gamma}{1-\gamma}\norm{a'-a''}_2,\label{eq:grad of Q 3}
\end{align}
where (\ref{eq:grad of Q 2}) follows from (\ref{eq:grad of Q as grad of r + grad of V}) and the triangle inequality, and (\ref{eq:grad of Q 3}) holds by the regularity assumptions of $r$ and $p$ as well as the fact that $\lvert V(s';\pi)\rvert\leq \frac{1}{1-\gamma}$.
This implies that $Q(s,a;\pi)$ is $L_Q$-smooth in $a$.

\underline{\textbf{For (ii):}} We adapt the proof technique in \citep{agarwal2019theory,agarwal2020optimality} and show the smoothness result of $V(s;\pi_{\theta})$ in $\theta$. 
Recall that by the Bellman equation, under a deterministic policy, we have 
\begin{equation}
        V(s;\pi)=r(s,\pi(s))+\gamma \sum_{s'\in\cS}p(s'\rvert s,a)V(s';\pi), \hspace{6pt}\forall s\in\cS.\label{eq:V as r+V}
\end{equation}
Let $r^{\pi}$ be an $M$-dimensional column vector where the $i$-th entry is $r(i,\pi(i))$, and $V^{\pi}$ denote an $M$-dimensional column vector where the $i$-th entry is $V(i;\pi)$. 
Then, we can rewrite (\ref{eq:V as r+V}) in matrix form as
\begin{equation}
    V^{\pi}=r^{\pi}+\gamma \bP^{\pi}V^{\pi}.\label{eq:V as r+V in matrix form}
\end{equation}
By (\ref{eq:V as r+V in matrix form}), we immediately know that 
\begin{align}
    V^{\pi}=(\bI_M-\gamma \bP^{\pi})^{-1} r^{\pi}.
\end{align}
For each $s\in\{1,\cdots,M\}$, let $e_s$ denote an $M$-dimensional one-hot vector with the $s$-th entry equal to $1$. Hence, for each $s\in\cS$, we know $V(s;\pi)=e_s^{\top}(\bI_M-\gamma \bP^{\pi})^{-1} r^{\pi}$. 
By Lemma \ref{lemma:matrix derivative}, we have
\begin{align}
    \frac{d V_s(\delta)}{d\delta}&=\gamma e_s^{\top} \bG(\delta)\frac{d \bP(\delta)}{d \delta}\bG(\delta)r^{\pi},\label{eq:1st-order derivative of V}\\
    \frac{d^2 V_s(\delta)}{d\delta^2}&=2 \gamma e_s^{\top} \bG(\delta)\frac{d \bP(\delta)}{d \delta}\bG(\delta)\frac{d \bP(\delta)}{d \delta}\bG(\delta) r^{\pi} +e_s^{\top} \bG(\delta)\frac{d^2 \bP(\delta)}{d \delta^2}\bG(\delta)r^{\pi}. \label{eq:2nd-order derivative of V}
\end{align}
Then, for any $\bW \in \bR^{M\times N}$ with $\norm{\bW}_F=1$, we have
\begin{align}
    \Big\lvert \frac{d V_s(\delta)}{d\delta}\Big\rvert &\leq \gamma \norm{\bG(\delta)\frac{d \bP(\delta)}{d \delta}\bG(\delta)r^{\pi}}_{\infty}\label{eq:norm bound of derivaitve of V 1}\\
    &\leq \frac{\gamma C_1}{(1-\gamma)^2},\label{eq:norm bound of derivaitve of V 2}
\end{align}
where (\ref{eq:norm bound of derivaitve of V 1}) is a direct result of (\ref{eq:1st-order derivative of V}), and (\ref{eq:norm bound of derivaitve of V 2}) follows from Lemmas \ref{lemma:infinite norm of G(z)x}-\ref{lemma:norm bound of derivative of P} and $\norm{r^{\pi}}_{\infty}\leq 1$.
Similarly, we know
\begin{align}
    \Big\lvert \frac{d^2 V_s(\delta)}{d\delta^2}\Big\rvert &\leq 2\gamma \norm{\bG(\delta)\frac{d \bP(\delta)}{d \delta}\bG(\delta)\frac{d \bP(\delta)}{d \delta}\bG(\delta) r^{\pi}}_{\infty}+ \norm{\bG(\delta)\frac{d^2 \bP(\delta)}{d \delta^2}\bG(\delta)r^{\pi}}_{\infty}\label{eq:norm bound of 2nd derivaitve of V 1}\\
    &\leq \frac{2\gamma C_1^2}{(1-\gamma)^3}+\frac{ C_2}{(1-\gamma)^2},\label{eq:norm bound of 2nd derivaitve of V 2}
\end{align}
where (\ref{eq:norm bound of 2nd derivaitve of V 1}) is a direct result of (\ref{eq:2nd-order derivative of V}), and (\ref{eq:norm bound of 2nd derivaitve of V 2}) holds by Lemmas \ref{lemma:infinite norm of G(z)x}-\ref{lemma:norm bound of derivative of P} and $\norm{r^{\pi}}_{\infty}\leq 1$.
Then, since (\ref{eq:norm bound of 2nd derivaitve of V 1})-(\ref{eq:norm bound of 2nd derivaitve of V 2}) hold for any $\bW \in \bR^{M\times N}$ with $\norm{\bW}_F=1$, then we know for every state $s$, $V(s;\pi_{\theta})$ is $L$-smooth in $\theta$, where $L=\frac{2\gamma C_1^2}{(1-\gamma)^3}+\frac{ C_2}{(1-\gamma)^2}$.
\MYQED
\end{myproof}

Now, we are ready to prove Proposition \ref{PROP: L-SMOOTH OF J}. For convenience, we restate Proposition \ref{PROP: L-SMOOTH OF J} below.
\begin{propstar}
Under the regularity assumptions (A1)-(A2), there exists some constant $L>0$ such that for any restarting state distribution $\mu$, $J_\mu(\pi(\cdot;\theta))$ is $L$-smooth in $\theta$.
\end{propstar}

\begin{myproof}[Proposition \ref{PROP: L-SMOOTH OF J}]
\normalfont By (ii) in Lemma \ref{lemma:Q and V L-smooth} and the definition that $J_{\mu}(\pi_\theta)=\E_{s\sim \mu}[V(s;\pi_\theta)]$, we know $J_{\mu}(\pi_\theta)$ is $L$-smooth, for any restarting state distribution $\mu$. \MYQED
\end{myproof}
\subsection{Proof of Proposition \ref{PROP:CONVERGENCE}}
\label{section:appendix:convergence proof}
For convenience, we restate Proposition \ref{PROP:CONVERGENCE} as follows.
\begin{propstar}
Under the FWPO algorithm with $\alpha_k(s)=\frac{(1-\gamma)\mu_{\min}}{L D_s^2}g_k(s)$, $\{\pi(\cdot;\theta_k)\}$ form a non-decreasing sequence of policies in the sense that $\pi(\cdot;\theta_{k+1})\geq \pi(\cdot;\theta_k)$, for all $k$.
Moreover, the effective Frank-Wolfe gap of FWPO converges to zero as $k\rightarrow \infty$, and the convergence rate can be quantified as
\begin{equation}
    \sum_{k=0}^{\infty} \cG_k^2\leq \frac{2LD_{\max}^2}{(1-\gamma)^3 \mu_{\min}^2},
\end{equation}
which implies that $\bar{\cG}_T=O(T^{-1/2})$.
\end{propstar}

To show the non-decreasing property in Proposition \ref{PROP:CONVERGENCE}, we summarize useful properties on policy improvement as follows.
We use $A(\cdot,\cdot;\pi)$ to denote the advantage function of a policy $\pi$.
\begin{lemma}[Performance difference lemma, \citep{kakade2002approximately}]
\label{lemma:performance difference}
For any two policies $\pi$ and $\pi'$, for any restarting state distribution $\mu$, the performance difference between the two policies at any state $s$ is
\begin{equation}
    V(s;\pi')-V(s;\pi)=\frac{1}{1-\gamma}\E_{s\sim d_{\mu}^{\pi'},a\sim\pi'(\cdot|s)}\big[A(s,a;\pi)\big].
\end{equation}
\end{lemma}
By Lemma \ref{lemma:performance difference}, we can directly obtain a sufficient condition of state-wise policy improvement.
\begin{corollary}
\label{corollary:state-wise policy improvement}
For any two policies $\pi$ and $\pi'$, we have $\pi'\geq \pi$ if the following condition holds for every state $s\in \cS$:
\begin{equation}
    \E_{a\sim \pi'}[A(s,a;\pi)]\geq \E_{a\sim \pi}[A(s,a;\pi)]=0.
\end{equation}
\end{corollary}

Now we are ready to prove Proposition \ref{PROP:CONVERGENCE}.

\begin{myproof}[Proposition \ref{PROP:CONVERGENCE}]
\normalfont For ease of notation, in this proof we use $\pi_k\equiv \pi(\cdot;\theta_k)$ and $\pi_k({s})\equiv \pi({s};\theta_k)$.
Recall from (\ref{eq:state-wise FW gap}) that under the policy $\pi_k$, the state-wise Frank-Wolfe gap is defined as $g_{k}(s):=\inp{c_k(s)-\theta_{k}(s)}{\nabla_{{a}} Q({s},{a};\pi_k)\rvert_{{a}=\pi_{k}({s})}}$.
By Lemma \ref{lemma:Q and V L-smooth}, the $Q(s,a;\pi)$ is $L$-smooth in $a$, for any state $s$ and any policy $\pi$. Then, under FWPO, we have
\begin{align}
    Q({s},\pi_{k+1}({s});\pi_k)&\geq Q({s},\pi_{k}({s});\pi_k)+\alpha_{k}(s)\inp{{c}_{k}(s)-\theta_{k}(s)}{\nabla_{{a}} Q({s},{a};\pi_k)\rvert_{{a}=\pi_{k}({s})}}-\frac{L}{2}\alpha_{k}(s)^2\norm{{c}_{k}(s)-\theta_{k}(s)}_2^2,\label{eq:Q improvement 1}\\
    &\geq Q({s},\pi_{k}({s});\pi_k)+{\alpha_{k}(s)g_{k}(s)-\frac{L}{2}\alpha_{k}(s)^2 D_{s}^2},\label{eq:Q improvement 2}
\end{align}
where (\ref{eq:Q improvement 2}) follows from the definitions of the state-wise Frank-Wolfe gap and the diameter $D_s$.
It is easy to verify that ${\alpha_{k}(s)g_{k}(s)-\frac{L}{2}\alpha_{k}(s)^2 D_{s}^2}$ is positive for all $\alpha_{k}(s)\in (0,\frac{2g_{k}(s)}{LD_s^2})$ and attains a maximum of $\frac{g_{k}(s)^2}{2L D_s^2}$ at $\alpha_{k}(s)=\frac{g_{k}(s)}{LD_s^2}$.
Therefore, if the learning rate $\alpha_{k}(s)\in(0,\frac{2g_{k}(s)}{LD_s^2})$, then $Q({s},\pi_{k+1}({s});\pi_k)>Q({s},\pi_{k}({s});\pi_k)$ and hence
$A({s},\pi_{k+1}({s});\pi_k)>0$.
By Corollary \ref{corollary:state-wise policy improvement}, we know $\pi_{k+1}\geq \pi_{k}$, for all $k$. 
Hence, $\{\pi(\cdot;\theta_k)\}$ indeed form a non-decreasing sequence of policies.

Next, we characterize the rate of convergence of the objective $J_{\mu}(\pi_{k})$ of the proposed FWPO algorithm.
By Proposition \ref{PROP: L-SMOOTH OF J}, we know the objective $J_{\mu}(\pi_{k})$ is $L$-smooth.
Therefore, we have
\begin{align}
    J_{\mu}(\pi_{k+1})&\geq J_{\mu}(\pi_{k})+\inp{\nabla_{\theta} J_{\mu}(\pi_k)}{\theta_{k+1}-\theta_{k}}-\frac{L}{2}\norm{\theta_{k+1}-\theta_{k}}_2^2 \label{eq:Jmu inequality 1}\\
    &=J_{\mu}(\pi_{k})+\sum_{s\in\cS}d^{\pi_k}_{\mu}(s)\alpha_{k}(s)\cdot\inp{{c}_{k}(s)-\theta_{k}(s)}{\nabla_{{a}} Q({s},{a};\pi_k)\rvert_{{a}=\pi_{k}({s})}}-\frac{L}{2}\sum_{s\in\cS}\alpha_{k}(s)^2\norm{{c}_{k}(s)-\theta_{k}(s)}_2^2 \label{eq:Jmu inequality 2}\\
    &\geq J_{\mu}(\pi_{k})+(1-\gamma)\mu_{\min}\sum_{s\in\cS}\alpha_{k}(s)g_{k}(s)-\frac{L}{2}\sum_{s\in\cS}\alpha_{k}(s)^2 D_s^2, \label{eq:Jmu inequality 3}
\end{align}
where (\ref{eq:Jmu inequality 1}) follows from that $J_{\mu}(\pi_{k})$ is $L$-smooth, (\ref{eq:Jmu inequality 2}) holds by the update scheme of FWPO, and (\ref{eq:Jmu inequality 3}) follows from that $d^{\pi_k}_{\mu}(s)\alpha_{k}(s)\geq (1-\gamma)\mu_{\min}$ and the definition of the diameters.
By using (\ref{eq:Jmu inequality 3}) and letting $\alpha_k(s)=\frac{g_k(s)}{L D_s^2}(1-\gamma)\mu_{\min}$,
\begin{align}
    J_{\mu}(\pi_{k+1})&\geq J_{\mu}(\pi_{k})+\sum_{s\in\cS}\frac{g_k(s)^2}{2L D_s^2}(1-\gamma)^2 \mu_{\min}^2 \label{eq:Jmu inequality 4}\\
    &\geq J_{\mu}(\pi_{k})+\frac{(1-\gamma)^2 \mu_{\min}^2}{2LD_{\max}^2}\sum_{s\in\cS}g_k(s)^2.\label{eq:Jmu inequality 5}
\end{align}
Recall that $\cG_k^2:=\sum_{s\in\cS}g_k(s)^2$ denotes the effective Frank-Wolfe gap of the $k$-th iteration.
By taking the telescoping sum of (\ref{eq:Jmu inequality 5}), we know
\begin{equation}
     J_{\mu}(\pi_{k+1})\geq  J_{\mu}(\pi_{0})+ \frac{(1-\gamma)^2 \mu_{\min}^2}{2LD_{\max}^2} \sum_{t=0}^{k} \cG_t^2.
\end{equation}
Let $\pi^*$ denote an optimal policy, i.e., $\pi^*\geq \pi$, for any policy $\pi$.
Hence, we have
\begin{equation}
    \sum_{t=0}^{k} \cG_t^2 \leq \frac{2LD_{\max}^2}{(1-\gamma)^2 \mu_{\min}^2}(J_{\mu}(\pi_{k+1})-J_{\mu}(\pi_{0}))\leq \frac{2LD_{\max}^2}{(1-\gamma)^2 \mu_{\min}^2}(J_{\mu}(\pi^*)-J_{\mu}(\pi_{0}))\leq \frac{2LD_{\max}^2}{(1-\gamma)^3 \mu_{\min}^2},\label{eq:bound on sum of effective FW gap}
\end{equation}
where the last inequality follows from the fact that the value functions are upper bounded by $(1-\gamma)^{-1}$.
Recall that $\bar{\cG}_T:=\min_{0\leq k\leq T}\cG_k$. 
Therefore, (\ref{eq:bound on sum of effective FW gap}) implies that
\begin{equation}
    \bar{\cG}_T\leq \sqrt{\frac{1}{T+1}\sum_{t=0}^{T} \cG_t^2}\leq \sqrt{\frac{1}{T+1}\cdot\frac{2LD_{\max}^2}{(1-\gamma)^3 \mu_{\min}^2}}=O(T^{-1/2}).
\end{equation}
This completes the proof. \MYQED
\end{myproof}

\section{Proof of Proposition \ref{PROP:NFWPO DDPG}}
\label{section:appendix:equivalence}
For convenience, we restate Proposition \ref{PROP:NFWPO DDPG} as follows.
\begin{propstar}
\label{PROP:NFWPO and DDPG}
If there is no action constraints, then the policy update scheme of NFWPO is equivalent to the vanilla DDPG by \citep{lillicrap2016continuous}.
\end{propstar}
\begin{myproof}
\normalfont We prove this result by reinterpreting the DDPG from a state-wise perspective.
Let $\bar{\theta}$ and $\bar{\phi}$ be the current parameters of the actor and the critic, respectively. 
As proposed in \citep{lillicrap2016continuous}, the policy update under DDPG is done by approximating the true gradient $\nabla_{\theta}J_\mu(\pi(\cdot;{\theta}))$ by the sample-based gradient $\nabla_{\theta}\hat{J}_\mu(\pi(\cdot;{\theta}))\approx \nabla_{\theta}J_\mu(\pi(\cdot;{\theta}))$ as
\begin{align}
    &\nabla_{\theta} \hat{J}_\mu(\pi(\cdot;{\theta}))=\frac{1}{\lvert \cB\rvert}\sum_{s\in\cB}\nabla_{a}Q(s,a; \phi)\rvert_{a=\pi(s;\bar{\theta})}\nabla_{\theta}\pi(s;\theta),\label{eq:sample-based DPG}
\end{align}
where $\cB$ denotes a mini-batch of states drawn from the replay buffer. 
Note that this update rule can be reinterpreted from the perspective of regression by the following steps: 
\vspace{-2mm}

\begin{itemize}[leftmargin=*]
    \item \textbf{Reference actions.} For each $s$ in the mini-batch $\cB$, compute the reference action at state $s$ under the guidance of the critic
    \begin{equation}
        a^{*}_s=\pi(s;\bar{\theta})+\eta_{1} \nabla_a Q(s,a;\bar{\phi})\rvert_{a=\pi(s;\bar{\theta})},\label{eq:DDPG reference action}
    \end{equation}
    where $\eta_1>0$ is the step size.
    \vspace{-1mm}
    \item \textbf{Loss function.} Construct a loss function $L(\theta;\bar{\theta})$ as the mean-squared error (MSE) between the actions of the current policy and the reference actions, i.e.,
    \begin{align}
        &\cL(\theta;\bar{\theta})=\frac{1}{2\lvert \cB\rvert}\sum_{s\in\cB}\big(\pi(s;\theta)- a^{*}_s\big)^2.\label{eq:DDPG MSE loss}
    \end{align}
    \vspace{-2mm}
    \item \textbf{Gradient update.} Accordingly, update the policy parameter by minimizing the MSE loss by applying gradient descent for one step, i.e.,
    \begin{equation}
        \theta\leftarrow\theta - \eta_{2} \nabla_\theta \cL(\theta;\bar{\theta}).\label{eq:DDPG MSE update}
    \end{equation}
\end{itemize}
We can observe that the update scheme characterized by (\ref{eq:DDPG MSE loss})-(\ref{eq:DDPG MSE update}) is equivalent to the DDPG update with a learning rate of $\eta_1\eta_2$.
Therefore, DDPG can be viewed as an actor-critic algorithm where both the actor and critic are trained by using regression as subroutines.
Note that (\ref{eq:DDPG reference action}) is equivalent to (\ref{eq:FW target action}) if there is no action constraints.
Hence, we conclude that NFWPO is equivalent to DDPG if there is no action constraints. \MYQED

\end{myproof}

\vspace{2mm}
\section{Pseudo Code of NFWPO}
\label{section:appendix:pseudo}
For completeness, we provide the pseudo code of NFWPO in Algorithm \ref{alg:NFWPO}.

\begin{algorithm}[!htbp]
   \caption{Frank-Wolfe Policy Optimization With Neural Policy Parameterization (NFWPO)}
   \label{alg:NFWPO}
\begin{algorithmic}[1]
   \STATE {\bfseries Input:} Frank-Wolfe learning rate $\alpha$, actor learning rate $\beta$, critic learning rate $\beta_c$, target update ratio $\tau$, and variance of exploration noise $\sigma^2$
   \STATE Randomly initialize the actor $\pi(\cdot;\theta)$ and the critic $Q(\cdot,\cdot;\phi)$ with parameters $\theta,\phi$
   \STATE Initialize the target networks with parameters $\theta^{\dagger},\phi^{\dagger}$
    \FOR{each episode $i=0,1,\cdots$}
      \STATE Let $\bar{\theta}$ and $\bar{\phi}$ be the snapshots of the current actor and critic parameters
      \STATE Receive initial state $s_0$ of the current episode
      \FOR{$t=0,1,\cdots$}
      \STATE Select action $a_t=\pi(s_t;\bar{\theta})+\mathcal{N}(0,\sigma^2)$ 
      \STATE Apply action $a_t$ and observe reward $r_t$ as well as the next state $s_{t+1}$
      \STATE Store transition $(s_t,a_t,r_t,s_{t+1})$ in the replay buffer
      \STATE Randomly sample a mini-batch $\cB$ of transitions $(s,a,r,s')$ from the replay buffer
      \STATE // update the actor by state-wise Frank-Wolfe
      \FOR{each state $s\in\cB$}
        \STATE $\bar{c}(s)=\argmax_{c \in \cC(s)}\inp{c}{\nabla_{a} Q(s,a;\bar{\phi})\rvert_{a=\Pi_{\cC(s)}(\pi(s;\bar{\theta}))}}$
        \STATE $\tilde{a}_s=\Pi_{\cC(s)}(\pi(s;\bar{\theta}))+{\alpha} \big(\bar{c}(s)-\Pi_{\cC(s)}(\pi(s;\bar{\theta}))\big)$
      \ENDFOR
        \STATE      $\cL_{\text{NFWPO}}(\theta;\bar{\theta})=\sum_{s\in\cB}\big(\pi(s;\theta)- \tilde{a}_s\big)^2$
        \STATE $\theta\leftarrow\theta - {\beta}\nabla_\theta \cL_{\text{NFWPO}}(\theta;\bar{\theta})\rvert_{\theta=\bar{\theta}}$
      \STATE // update the Q-learning-like critic as vanilla DDPG
      \STATE $\cL_{\text{critic}}(\phi)=\frac{1}{\lvert \cB\rvert}\sum_{(s,a,r,s')\in\cB}\big(r+\gamma Q(s',\pi(s';\theta^\dagger);\phi^\dagger)-Q(s,a;\phi)\big)^2$
      \STATE $\phi\leftarrow \phi - \beta_c \nabla_{\phi}\cL_{\text{critic}}(\phi)\rvert_{\phi=\bar{\phi}}$
      \STATE Update the target networks: $\theta^\dagger\leftarrow \tau \theta+(1-\tau)\theta^\dagger$, $\phi^\dagger\leftarrow \tau \phi+(1-\tau)\phi^\dagger$
      \ENDFOR
    \ENDFOR
\end{algorithmic}
\end{algorithm}

\section{Detailed Experimental Setup and Additional Experimental Results}
\label{section:appendix:exp}
\subsection{Training Configurations}
\textbf{Random seeds}. For each task, each algorithm is trained under the common set of random seeds of $\{0,1,2,3,4\}$. 

\textbf{Exploration.} The training process starts after some number of time steps (1000 steps for Reacher-v2 and 10000 steps for the other environments), and we use a purely exploratory policy in this initial phase to collect samples for the replay buffer for all the algorithms. During training, Gaussian noise is added to each action for exploration for the neural cases. In the tabular case, we use the $\epsilon$-greedy policy as the behavior policy instead. 

\textbf{Update frequency}. For NFSNET, due to the longer computation time required by the network simulator, we speed up the simulations by updating the actor every 50 steps for all the algorithms. 
For the other environments, as DDPG+OptLayer is particularly time-consuming, we also update its actor every 50 steps to speed up the training process.

\textbf{Implementation of OptLayer.} As there is no off-the-shelf implementation of DDPG+OptLayer available, we leverage the open-source packages cvxpylayer \citep{agrawal2019differentiable} as well as the OptNet proposed by \citep{amos2017optnet} to implement the differentiable projection-based OptLayer for end-to-end training.
Note that in DDPG+OptLayer there are two scenarios where projection is needed for the actor output: (i) producing actions for interacting with the environment (under a behavior policy) and (ii) evaluating the actions produced by the current policy for calculating the deterministic policy gradient as in (\ref{eq:sample-based DPG}).
Since OptLayer is computationally inefficient, we use it only for the scenario (ii), where back-propagation is required for gradient descent.
For scenario (i), we use Gurobi optimization solver instead to speed up the training process.

\subsection{A Summary of the Hyperparameters}
In this section, we provide a summary of the key hyperparameters in Tables \ref{tab:hyperParam neural}-\ref{tab:hyperParam tabular}.
We highlight some key design choices:
\begin{itemize}[leftmargin=*]
    \item For both the actor and critic networks, we use two hidden layers with 400 and 300 hidden units with ReLU as the activation, as suggested by \cite{fujimoto2018addressing}. For the activation function of the actor output, in order to better accommodate the various action ranges of different environments, we choose tanh for the MuJoCo control tasks and ReLU for the other tasks, respectively. 
    \item We choose a smaller batch size for DDPG+OptLayer since its computation time is much larger than the other approaches. As DDPG+OptLayer is a strong baseline in some of the environments, we also set the batch size of NFWPO to be 16 for a fair comparison.
    
    \item For SAC algorithm, we use the open-source Spinning up \citep{SpinningUp2018} and its default setting. For PPO and TRPO, we use the open-source provide by \citep{fujita2018clipped}, which is based on ChainerRL \citep{JMLR:v22:20-376} with its default setting.

\end{itemize}

\subsection{Additional Experimental Results}
In this section, we further compare NFWPO with the baseline methods designed to address special types of action constraints in the HalfCheetah-v2 environment.

\begin{figure*}[!htb]
\centering
$\begin{array}{c} 
    \multicolumn{1}{l}{\mbox{\bf }}\\ 
    \hspace{-1mm} \scalebox{0.3}{\includegraphics[width=\textwidth]{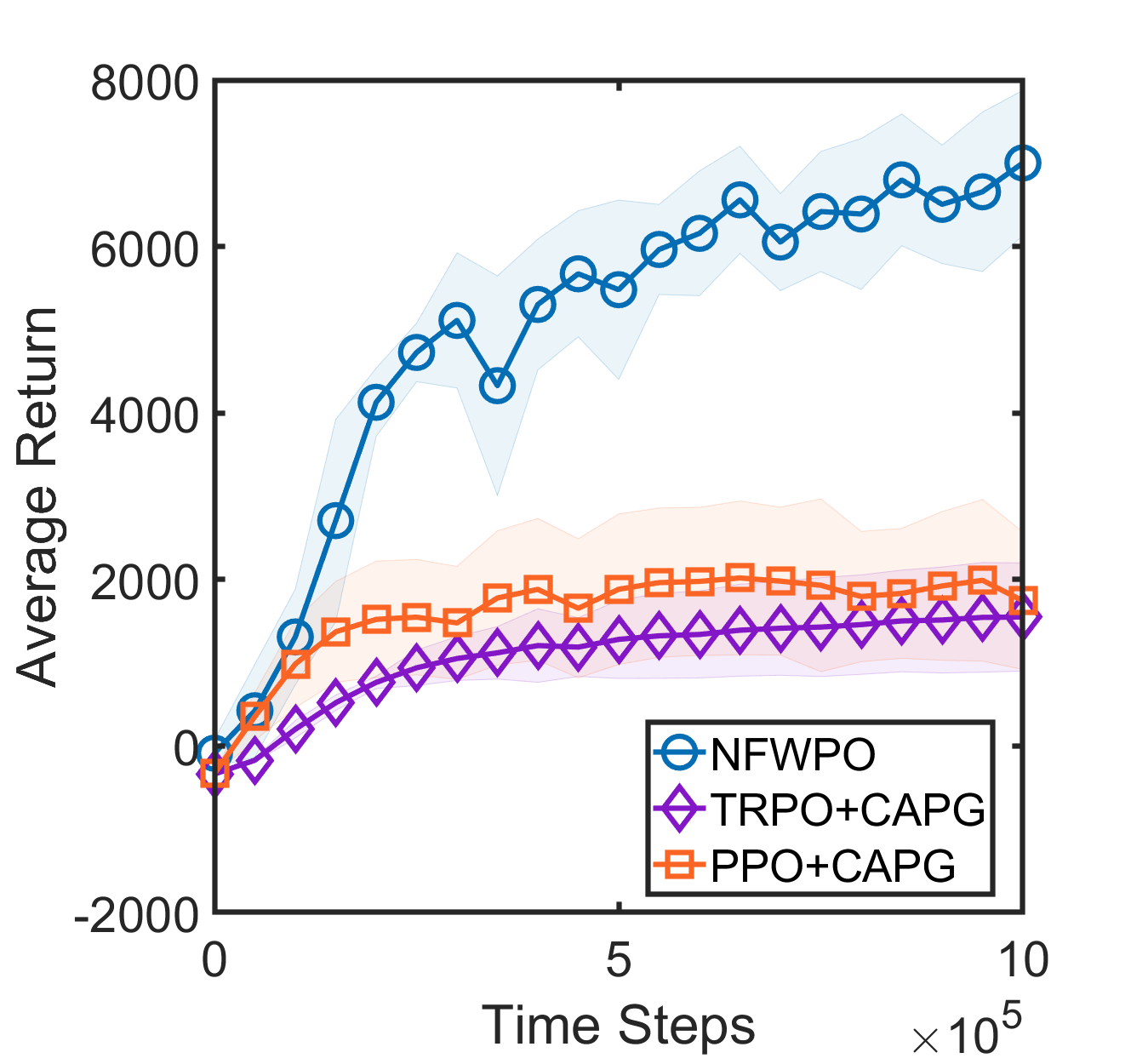}}  \label{fig:Halfcheetah_bound}
\end{array}$
\caption{Halfcheetah-v2 with bound constraints: Average return at each evaluation over 5 random seeds}
\label{fig:Halfcheetah_bound_all}
\end{figure*}

\textbf{Comparison with Clipped Action Policy Gradient \citep{fujita2018clipped}.} 
We further compare NFWPO with the Clipped Action Policy Gradient (CAPG) algorithms.
As CAPG can only handle bound constraints, we evaluate them with the intrinsic bound constraints in HalfCheetah (i.e., each action entry must be in $[-1,+1]$). 
The results in Figure \ref{fig:Halfcheetah_bound_all} show that NFWPO outperforms the two versions of CAPG, namely TRPO-CAPG and PPO-CAPG, under the bound constraints. 

\begin{table}[H]
    \centering
    \caption{A summary of the hyper-parameters of NFWPO and the baseline methods.}
    \label{tab:hyperParam neural}
    \begin{tabular}{ l  c c c c}
      \hline Hyper-parameters & Reacher-v2 & HalfCheetah-v2 & NSFNET & BSS-5\\
      \hline
      Learning Rate (For Frank-Wolfe) & 0.05 & 0.01 & 0.05 & 0.05 \\
      Actor Learning Rate (For others) & 0.0001 & 0.0001 & 0.0001 & 0.0001 \\
      Critic Learning Rate & 0.001 & 0.001 & 0.001 & 0.001\\
      Discount Factor & 0.99 & 0.99 & 0.99 & 0.99\\
      Target Update Ratio ($\tau$) & 0.001 & 0.001 & 0.001 & 0.001\\
      Replay Buffer Size & $10^4$ & $10^6$ & $5\times10^4$ & $10^6$\\
      Evaluation Frequency & 5000 & 5000 & 10000 & 5000\\
      Total Training Steps & $3\times10^5$ & $7\times10^5$ & $5\times10^5$ & 100000 \\
      Starting Time of Training & 1000 & 10000 & 10000 & 10000\\
      Additive Action Noise for Exploration & $\mathcal{N}(0, 0.1) $ & $\mathcal{N}(0, 0.1)$ & $\mathcal{N}(0, 3)$ & $\mathcal{N}(0, 5)$\\
      Weight of Reward Shaping  & $\frac{1}{7}$ &  3 (state-dependent), 2 (quadratic)& $\frac{1}{4}$ & 4\\
      Batch Size (For DDPG+RewardShaping) & 64         & 64             & 64 & 64\\
      Batch Size (For DDPG+Projection)& 64         & 64             & 64 & 64\\
      Batch Size (For DDPG+OptLayer)  & 16         & 16             & 16 & 16\\
      Batch Size (For FWPO/NFWPO)          & 16         & 16             & 16 & 16 \\
      \hline  
    \end{tabular}
\end{table}

\begin{table}[H]
    \centering
    \caption{The configurations of the hyper-parameters for tabular policies in the bike-sharing environment.}
    \label{tab:hyperParam tabular}
    \begin{tabular}{ l  c}
      \hline Hyper-parameters & BSS-3\\
      \hline
      Critic Network & (30, out) \\
      Learning Rate (For Frank-Wolfe) & 0.05 \\
      Actor Learning Rate (For Others) & 0.001 \\
      Critic Learning Rate & 0.002 \\
      Discount Factor & 0.9 \\
      Target Update Ratio for Critic ($\tau$) & 0.01 \\
      Target Update Frequency for Actor & 100 \\
      Replay Buffer Size & $10^4$ \\
      Evaluation Frequency & 5000 \\
      Total Training Steps & $10^6$ \\
      Starting Time of Training & 10000 \\
      Exploratory Behavior Policy & $\epsilon$-greedy, $\epsilon = 0.1$ \\
      Weight of Reward Shaping & 1\\
      Batch Size (For All Methods) & 64 \\
      \hline  
    \end{tabular}
\end{table}